\documentclass[10pt,twocolumn,letterpaper]{article}
\usepackage[T1]{fontenc} %

\usepackage{cvpr}

\usepackage{overpic}
\usepackage{amsmath}
\usepackage{amssymb}
\usepackage{algorithm}
\usepackage{listings}
\usepackage{array}
\usepackage{colortbl}
\definecolor{AppendixRowGray}{HTML}{F4F4F4}
\newcolumntype{P}[1]{>{\raggedright\arraybackslash}m{#1}}
\usepackage{iftex}
\ifPDFTeX
\fi
\usepackage{arydshln}
\usepackage{etoolbox}
\usepackage{pifont}
\usepackage{makecell}
\usepackage{multirow}
\usepackage{enumitem}
\usepackage{bm}
\usepackage{diagbox}
\usepackage{cancel}

\DeclareSymbolFont{bbold}{U}{bbold}{m}{n}
\SetSymbolFont{bbold}{bold}{U}{bbold}{bx}{n}

\makeatletter
\AfterEndEnvironment{algorithm}{\let\@algcomment\relax}
\AtEndEnvironment{algorithm}{\kern2pt\hrule\relax\vskip3pt\@algcomment}
\let\@algcomment\relax

\newcommand\algcomment[1]{\def\@algcomment{\footnotesize#1}}
\renewcommand\fs@ruled{\def\@fs@cfont{\bfseries}\let\@fs@capt\floatc@ruled
  \def\@fs@pre{\hrule height.8pt depth0pt \kern2pt}%
  \def\@fs@post{}%
  \def\@fs@mid{\kern2pt\hrule\kern2pt}%
  \let\@fs@iftopcapt\iftrue}
\makeatother

\newcommand{\tablestyle}[2]{\setlength{\tabcolsep}{#1}\renewcommand{\arraystretch}{#2}\centering\small}

\definecolor{new_olive}{HTML}{8b9656}

\newcommand{\correct}{{\small\color{new_olive}\ding{51}}}
\definecolor{new_pink}{HTML}{e096a7}

\newcommand{\wrong}{{\small\color{new_pink}\ding{56}}}

\definecolor{cvprblue}{rgb}{0.21,0.49,0.74}
\usepackage[pagebackref,breaklinks,colorlinks,allcolors=cvprblue]{hyperref}
\AtEndPreamble{%
  \crefname{figure}{Fig.}{Figs.}%
  \Crefname{figure}{Fig.}{Figs.}%
  \crefname{table}{Tab.}{Tabs.}%
  \Crefname{table}{Tab.}{Tabs.}%
  \crefname{section}{Sec.}{Secs.}%
  \Crefname{section}{Sec.}{Secs.}%
  \crefname{subsection}{Sec.}{Secs.}%
  \Crefname{subsection}{Sec.}{Secs.}%
  \crefname{subsubsection}{Sec.}{Secs.}%
  \Crefname{subsubsection}{Sec.}{Secs.}%
}

\graphicspath{{./figs/}}
\DeclareGraphicsExtensions{.pdf,.jpg,.png}

\newcommand{\myPara}[1]{\vspace{2pt}\noindent\textbf{#1}}

\def\confName{CVPR}
\def\confYear{2027}
\definecolor{lightgreen}{RGB}{64,128,128}

\title{ Game408: Deep Reasoning from Clues to Actions in \\ Long-Horizon Narrative Game Worlds \vspace{-10pt}}

\author{
  Jing Wang\textsuperscript{1 *}\qquad
  Modi Jin\textsuperscript{1 *}\qquad
  Shichao Chen\textsuperscript{2 \textdagger}\\[5pt]
  \textsuperscript{1}Nankai University \quad
  \textsuperscript{2}Institute of Automation, Chinese Academy of Sciences\\[3pt]
  {\textsuperscript{*}Equal contribution.\qquad
   \textsuperscript{\textdagger}Corresponding author.}\\[3pt]
  \textit{\{Wjing, jin\_modi\}@mail.nankai.edu.cn\qquad shichao.chen@ia.ac.cn}
}
\AtBeginDocument{\hypersetup{pdfauthor={Jing Wang, Modi Jin, Shichao Chen}}}
\begin{document}
\twocolumn[{%
\renewcommand\twocolumn[1][]{#1}%
\maketitle
\vspace{-1.2cm}
\begin{center}
    \centering
    \setlength{\abovecaptionskip}{2pt}
    \includegraphics[width=\textwidth]{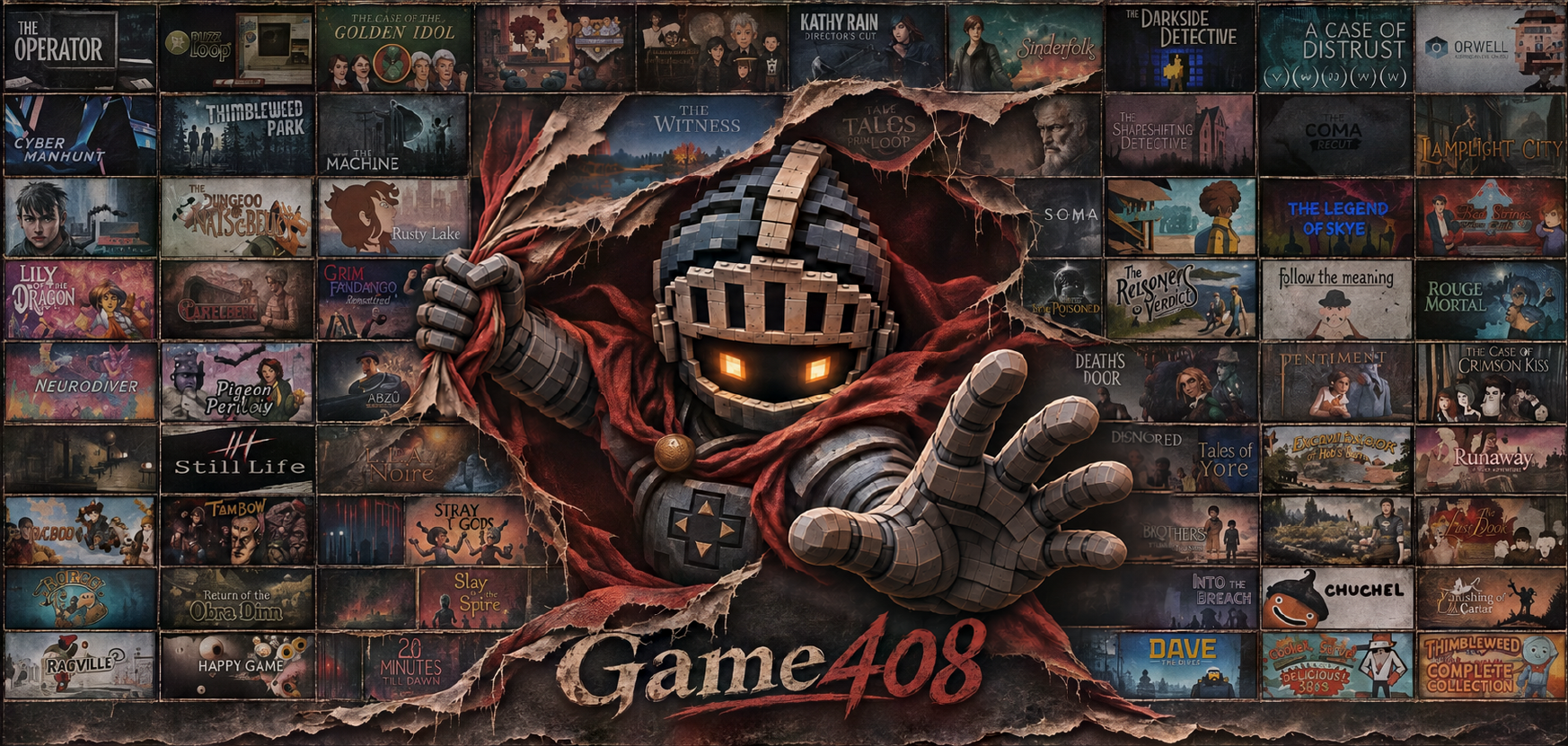}
    \captionsetup{hypcap=false}
    \captionof{figure}{
\textbf{Game408 dataset.}
It includes 54 games, 408 checkpoints, and 24K+ human reasoning and action annotations.
}
    \captionsetup{hypcap=true}
    \label{fig:teaser}
\end{center}%
}]

\begin{abstract}
\renewcommand{\endabstract}{\par}
Long-horizon narrative reasoning games require agents to integrate clues, retrieve memories, acquire evidence and advance the story through concrete actions.
Reasoning and execution are interdependent during interaction, while a shared outcome signal makes it difficult to distinguish errors in intent from failures in execution.
To address this problem, we construct Game408, a dataset for game reasoning that provides fine-grained supervision across 54 games, 408 checkpoints, and over 24K human reasoning and action records.
Building on Game408, we introduce GameHunter.
Supervised fine-tuning establishes reasoning and action generation, followed by stage-wise decoupling through action bootstrapping to focus on execution.
Local distillation on autonomous trajectories then jointly trains reasoning and actions for sustained coordination.
Experiments show that GameHunter outperforms general-purpose models and existing game agents in both task completion rate and action selection accuracy on long-horizon reasoning games.
Ablations validate the effectiveness of stage‑wise decoupling and local distillation for autonomous trajectory generation.
\end{abstract}

\section{Introduction}
\label{sec:intro}

Narrative reasoning games require agents to turn clues into executable decisions throughout ongoing interaction~\cite{hausknecht2020jericho,ruan2025flashadventure,cote2018textworld,qian2025escapebench,ammanabrolu2020kga2c}.
Evidence discovered early may only become useful several scenes later, while each action determines what information can be acquired next~\cite{ruan2025flashadventure,lim2025visescape,cote2018textworld}.
Agents must connect evidence, retrieve memories, and revise their judgments while accurately selecting targets, tools, and control parameters, so that reasoning can continue to advance the story~\cite{lim2025visescape,tan2024cradle,li2024optimus1,qin2024mp5}.

Prior work has improved game-playing capabilities through reinforcement learning, behavioral learning, language-based planning, and external memory~\cite{mnih2015human,silver2016mastering,vinyals2019grandmaster,baker2022video,wang2023voyager,wang2023jarvis1}.
MineRL~\cite{guss2019minerl} and VPT~\cite{baker2022video} learn control from human gameplay.
Voyager~\cite{wang2023voyager}, Cradle~\cite{tan2024cradle}, and JARVIS-1~\cite{wang2023jarvis1} use planning, tools, and memory for sustained exploration.
Related reasoning and reflection methods use interaction feedback to revise decisions~\cite{yao2023react,shinn2023reflexion,madaan2023selfrefine,wang2023deps,lee2026xenon,hu2024pokellmon}.
MineDojo~\cite{fan2022minedojo} introduces open-ended tasks and internet-scale knowledge.
Generalist and cross-game models further broaden coverage~\cite{reed2022generalist,wang2025gametars,magne2026nitrogen,zhang2026scaling,hafner2023dreamerv3,team2025sima2}.
Recent benchmarks also expand this coverage~\cite{zhang2026gameverse,wu2023smartplay,paglieri2025balrog,kim2025orak,zhang2025korgym,chen2024gamearena}.
FlashAdventure~\cite{ruan2025flashadventure} evaluates complete adventure story arcs, and Lumine~\cite{tan2025lumine} demonstrates story missions lasting several hours.
These advances provide a foundation for studying complex games.
However, detailed human annotations that link clues, earlier evidence, action purposes, and recovery processes step by step across diverse narrative games remain limited, as shown in \cref{tab:dataset}.

\begin{table}[t]
   \small
   \centering
   \setlength{\abovecaptionskip}{2pt}
   \caption{\textbf{Comparison of representative game datasets and environment resources.} Language, CoT, and narrative indicate the availability of language annotations, process-level supervision, and narrative gameplay, respectively. Games denotes the number of covered games.}
   \setlength{\abovecaptionskip}{2pt}
   \tablestyle{1.8pt}{1.0}
   \begin{tabular}{lcccc}
     \hline

     \hline

     \hline
     \textbf{Resource} & \textbf{Language}& \textbf{CoT}& \textbf{Narrative}& \textbf{Games} \\
     \midrule
     Atari Grand Challenge~\cite{kurin2017atari} & \wrong & \wrong & \wrong & 5 \\
     MineRL~\cite{guss2019minerl} & \wrong & \wrong & \wrong & 1 \\
     VPT dataset~\cite{baker2022video} & \wrong & \wrong & \wrong & 1 \\
     MineDojo~\cite{fan2022minedojo} & \correct & \wrong & \wrong & 1 \\
     Jericho~\cite{hausknecht2020jericho} & \correct & \wrong & \correct & 56 \\
     TALES~\cite{wu2025tales} & \correct & \wrong & \correct & 122 \\
     FlashAdventure~\cite{ruan2025flashadventure} & \correct & \wrong & \correct & 34 \\
     NitroGen dataset~\cite{magne2026nitrogen} & \wrong & \wrong & \wrong & 1{,}000+ \\
     AceAction-5M~\cite{ace2026action} & \correct & \wrong & \wrong & 42 \\
     GameHorizon-Data~\cite{wang2026gamehorizon} & \correct & \wrong & \wrong & 21 \\
     GameVerse~\cite{zhang2026gameverse} & \correct & \wrong & \wrong & 15 \\
     GameWorld~\cite{xu2026gameworld} & \correct & \wrong & \wrong & 34 \\
     VisEscape~\cite{lim2025visescape} & \correct & \wrong & \wrong & 20 \\
     Lumine reasoning data~\cite{tan2025lumine} & \correct & \correct & \correct & 1 \\
     \rowcolor[HTML]{fff5f4}
     \textbf{Game408 (Ours)} & \correct & \correct & \correct & \textbf{54} \\
      \hline

     \hline

     \hline
   \end{tabular}
   \label{tab:dataset}
\end{table}

The interdependence of reasoning and execution also makes credit assignment from final outcomes more difficult~\cite{arjona2019rudder,lightman2023verify,li2025optimus3,shi2026odysseus}.
Although action outputs are short, an incorrect target or parameter can block access to crucial information, leaving the agent to repeat attempts without making further inferences~\cite{lim2025visescape,long2026longpuzzle,qin2024mp5,lee2026xenon}.
When reasoning and actions are updated together using a trajectory-level outcome signal, a failed operation may penalize both correct reasoning and an incorrect action.
Conversely, a successful operation may reinforce both unsupported reasoning and an effective action.
Training therefore needs to distinguish whether an intent is reasonable from whether an operation fulfills that intent, and then learn their coordination across continuous interaction.

To this end, we first construct process data that preserve the basis of players' decisions and support distinguishing intent formation from action execution.
We build Game408 around the evidence linking successive decisions.
It covers 54 narrative puzzle games, with over 260 hours of human gameplay and 24K+ human reasoning and action samples.
Annotators connect current clues and earlier memories to action intent, and specify the target and function of each operation.
Exploratory and failed actions are retained together with recovery reasoning, allowing the model to learn how information supports judgments and how interaction feedback prompts players to revise their decisions.

To address the difficulty of distinguishing intent errors from execution errors under a shared outcome signal, we propose a strategy that decouples reasoning and execution across training stages.
GameHunter first learns to generate reasoning and actions from observations through supervised fine-tuning, then performs action bootstrapping conditioned on teacher reasoning.
At this stage, teacher reasoning fixes the current action intent, and the student focuses on translating it into concrete operations, concentrating training feedback on execution quality.
We evaluate whether action functions, targets, and control parameters match the intent, retain functionally equivalent implementations, and use advantage truncation to prevent incorrect actions from receiving positive training weight.

Reliable execution must then be integrated into the student's own reasoning process.
In the third training stage, GameHunter interacts autonomously and triggers local on-policy distillation based on recent decision quality, jointly supervising reasoning and actions on its actual trajectories.
The student can follow reasonable alternative exploration orders and receives corrections for the current context when quality declines.
Game state and history are retained after parameter updates, so that previously acquired evidence continues to support subsequent decisions.
Evaluation at story checkpoints further examines whether reasoning and operations can consistently support task completion, memory use, and error recovery.
Our contributions are:

\begin{itemize}[leftmargin=*,itemsep=1pt,topsep=2pt]
    \item We construct the Game408 dataset and benchmark, providing over 24K human annotations across games that link evidence, intent, operations, and recovery, together with evaluation at story checkpoints.

    \item We introduce GameHunter with a training strategy for long-horizon reasoning games that combines decoupling across stages with joint learning on autonomous trajectories.
The strategy focuses first on execution learning and then on joint distillation of reasoning and actions.

    \item We show that GameHunter outperforms existing baselines in task completion and action selection.
We also demonstrate through ablations the need to decouple reasoning and execution across stages.
\end{itemize}

\begin{figure*}[t]
  \centering
  \setlength{\abovecaptionskip}{2pt}
  \includegraphics[width=0.95\textwidth]{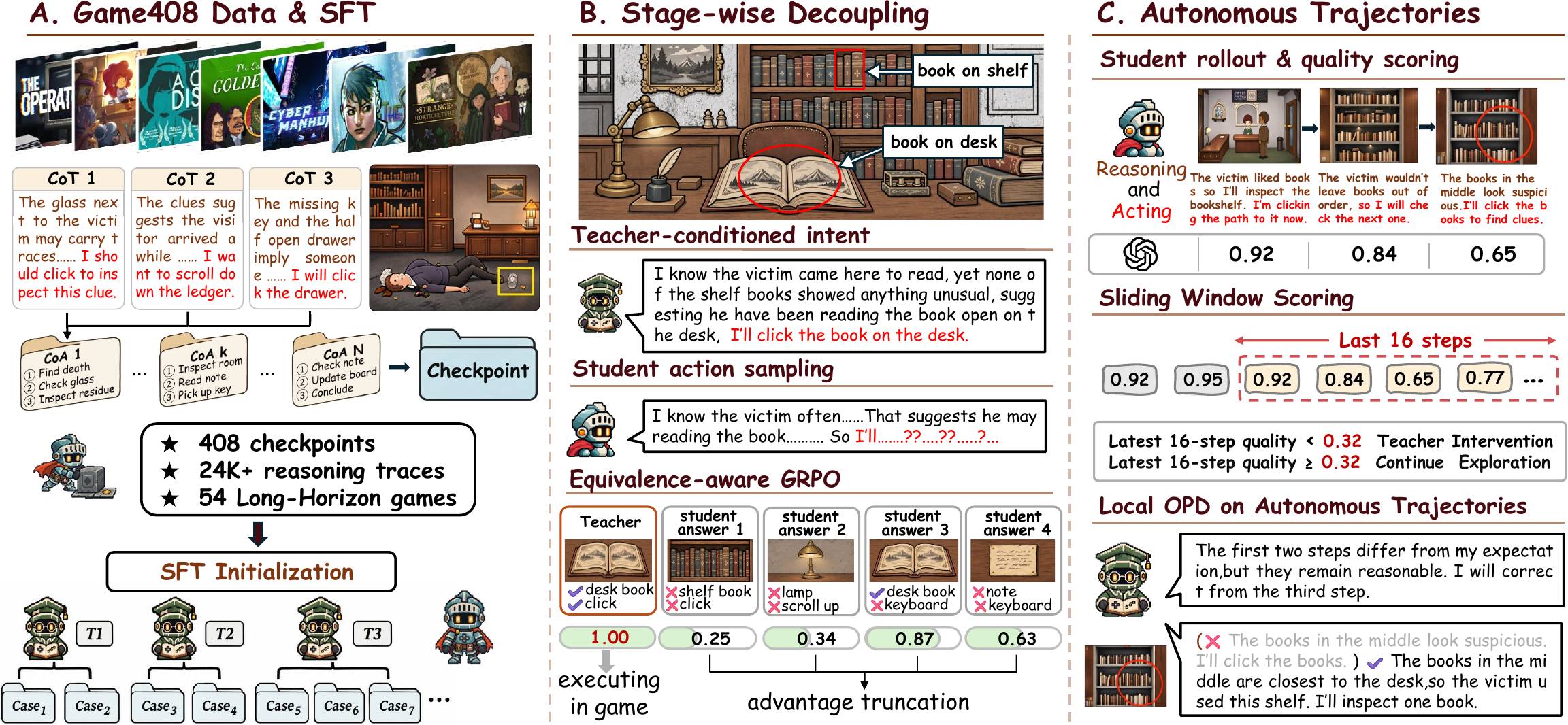}

\caption{\textbf{GameHunter training pipeline.}
(A) SFT uses Game408 annotations.
(B) Stage-wise decoupling trains execution with teacher-fixed intent and correctness-constrained GRPO.
(C) Trajectory quality triggers local joint OPD on autonomous student trajectories.}

  \label{fig:pipeline9} %
\end{figure*}

\section{Related Work}
\label{sec:relatedwork}

\subsection{Video Game Agent}

Game agents perceive and understand the game world to complete tasks through continuous interaction~\cite{bellemare2013ale,cobbe2020procgen,kuttler2020nle,schrittwieser2020muzero,ecoffet2021goexplore,narasimhan2015textgames}.
Early research mainly focuses on keyboard and mouse actions, learning game control through reward optimization or imitation of human operations~\cite{kurin2017atari,baker2022video,hester2018dqfd,li2023steve1,cai2023groot}.
Recent research has introduced LLMs and VLMs to support task planning and experience reuse through reasoning~\cite{wang2023jarvis1,li2024optimus1,ma2024textstarcraft,park2023generative,wang2023deps,qin2024mp5}.
This has gradually shifted game interaction from fixed steps toward adaptive exploration~\cite{wang2023voyager,tan2024cradle,zhang2026gameverse,lee2026xenon}.
JARVIS-VLA~\cite{li2025jarvis}, Game-TARS~\cite{wang2025gametars}, and NitroGen~\cite{magne2026nitrogen} learn game control from visual behavior data.
ROCKET-1~\cite{Cai_2025_CVPR} uses visual-temporal prompts to connect high-level reasoning with spatially grounded control.
Scaling in Games~\cite{zhang2026scaling} further examines the relationship between model scale and general game capabilities.
These studies provide a foundation for general game control~\cite{li2025optimus2,wang2024omnijarvis,wang2024minedreamer}.
However, visual--action correspondence alone leaves unclear how earlier evidence supports the current action intent.
Reasoning and execution also depend on each other in long-horizon games~\cite{wang2023voyager,tan2024cradle,li2024optimus1,wang2023deps,qin2024mp5,hu2024civrealm}.
When both rely on a shared outcome reward, it is difficult to distinguish intent errors from execution failures.
Execution learning for long-horizon narrative puzzle games therefore needs to specify the intent that an operation should fulfill.
Agents must then learn sustained coordination between reasoning and actions through autonomous interaction~\cite{li2025optimus3,shi2026odysseus}.

\subsection{Video Game Dataset}

Game datasets record observations, operations, and interaction feedback from players or agents~\cite{kurin2017atari,guss2019minerl,yao2020calm,karten2025pokechamp,hambro2022nld,li2025optimus2}.
MineRL~\cite{guss2019minerl} provides human state--action demonstrations for learning game control.
VPT~\cite{baker2022video} uses inverse dynamics to label actions in online gameplay videos.
MineDojo~\cite{fan2022minedojo} combines gameplay videos, tutorials, and internet knowledge with language-described tasks.
NitroGen~\cite{magne2026nitrogen} extracts frame-level actions from gameplay videos with input overlays across games.
Lumine~\cite{tan2025lumine} adds human reasoning at key decisions in a Genshin Impact story quest. Think-aloud data on ARC-AGI-3~\cite{liu2026howhumans} links human actions to mechanism discovery and error recovery.
These efforts enrich behavioral, knowledge, and reasoning supervision~\cite{chen2024gamearena,wang2024omnijarvis,li2025optimus3}. Complementary benchmarks examine general game capabilities~\cite{hu2026lmgamebench}, episodic memory~\cite{li-etal-2026-emembench}, narrative deduction~\cite{yuan-etal-2025-turnaboutllm}, and sustained text-game reasoning~\cite{phan2025textquests}.
However, long-horizon narrative puzzle solving also requires a continuous account of how current clues and earlier evidence shape action intent, and how interaction feedback changes subsequent judgments~\cite{ruan2025flashadventure,lim2025visescape,wu2025tales,qian2025escapebench}.
Game408 aligns human reasoning and actions across consecutive decisions, providing cross-game process supervision that connects evidence, intent, operations, and failure recovery.

\section{Game408 Dataset}

\begin{figure*}[t]
  \centering
  \setlength{\abovecaptionskip}{2pt}  \includegraphics[width=0.95\textwidth]{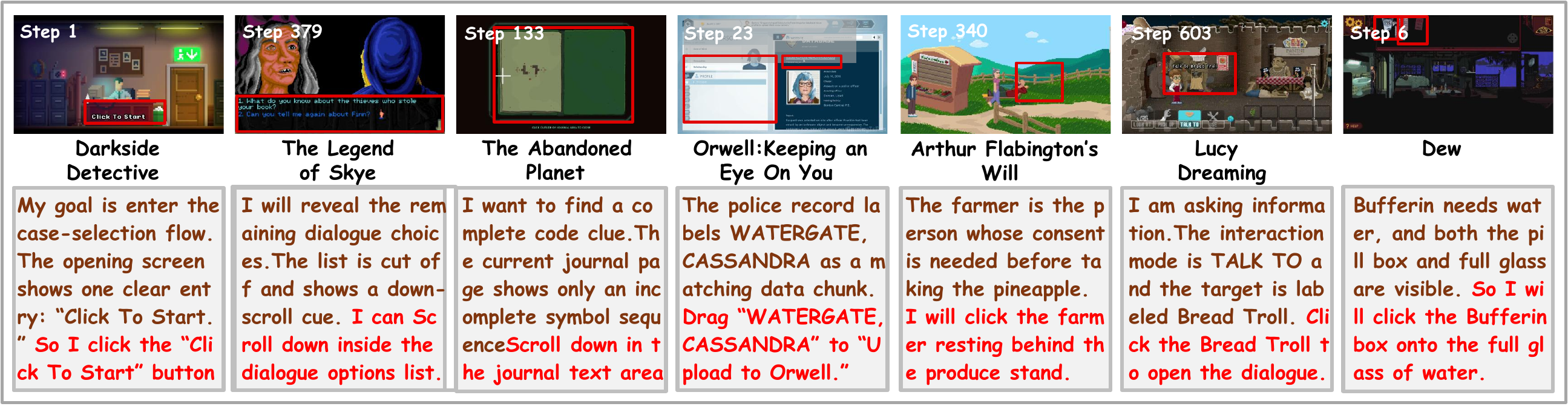}
\caption{\textbf{Game408-CoT examples from seven games.}
Each example links the observed scene and interaction target to human reasoning about the current state and the action, covering investigation, dialogue, clue retrieval, exploration, and object interaction.}

  \label{fig:Action3} %
\end{figure*}

Game408 provides over 24K human-annotated reasoning and action records for studying evidence use, intent formation, and action execution in long-horizon narratives.

\myPara{Game408-CoT. }Game408-CoT focuses on how each action arises from the reasoning process.
We collected over 260 hours of real human gameplay recordings and first used GPT-5.6 to coarsely parse the actions in the videos.
For each action, human annotators then added the connections among current clues, prior evidence, action intent, and interaction feedback, forming detailed chains of thought that connect adjacent decisions.
The annotations follow an observation--judgment--action--feedback sequence and are organized by story stage and checkpoint.
This organization allows models to learn how evidence supports intent, how intent constrains operations, and how feedback prompts subsequent corrections.
We standardize the action format and use GPT-4o to structure the CoT data into two parts: reasoning and actions.
\cref{fig:Action3} presents representative data from seven games.
These examples primarily show the model's reasoning about the current state and the corresponding actions.
Complete data annotations and descriptions are provided in the supplementary material.

\myPara{Game408-CoA. }Game408-CoA records continuous actions and error recovery.
We retain both efficient game-completion steps and actual errors and suboptimal exploration from human trajectories.
For recovery scenarios not covered by the original trajectories, we add actions such as retrying, switching tools, and repositioning based on the goals, available tools, and interaction feedback.
For repeated clicks, ineffective exploration, and prolonged stalls, we adjust action sequences and outcome labels and construct corresponding reflection steps.
Action chains are organized into checkpoints by story stage.
Expert-player volunteers handle story segmentation and difficulty balancing, while novices check whether tasks are too difficult or easy.

\myPara{Game408-Val. }
Game408 covers 54 games and 408 checkpoints.
We select 12 games and 83 checkpoints as the evaluation set, forming Game408-Val.
Both the data split and the evaluation data undergo manual verification to confirm that training and evaluation data do not overlap.
Evaluation is organized around the stage objectives associated with these checkpoints.
It assesses whether the model can complete tasks and sustain progress using clues and interaction feedback independently, and reports metrics including task completion rate, action quality, and reasoning quality.

\section{Methodology}

\begin{figure*}[t]
  \centering
  \setlength{\abovecaptionskip}{2pt}
  \includegraphics[width=0.95\textwidth]{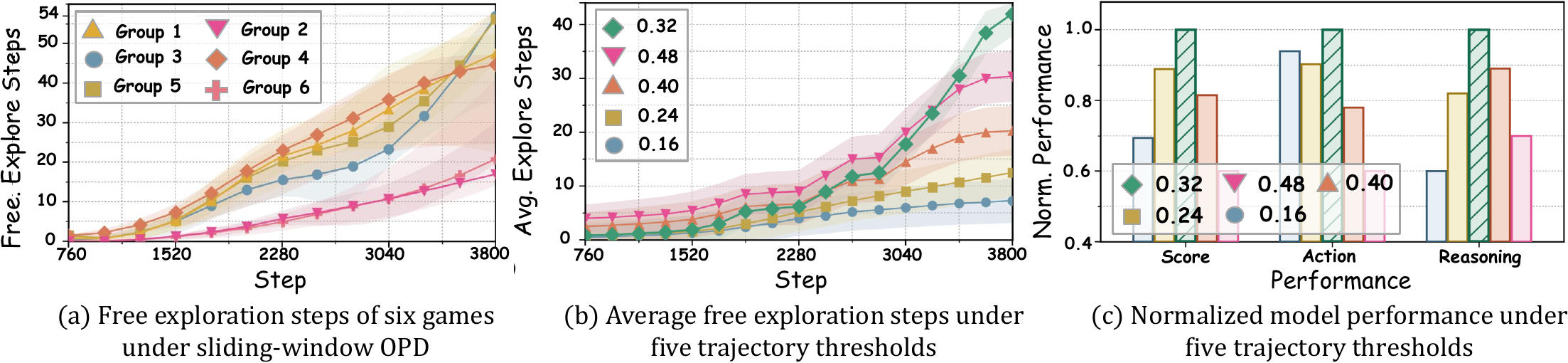}
\caption{\textbf{Sliding Window OPD threshold analysis.}
(a) Free-exploration steps for six groups of games.
(b) Mean exploration steps under five trajectory-quality thresholds.
(c) Normalized Score, Action, and Reasoning compare performance across thresholds.}

  \label{fig:32figures} %
\end{figure*}

We progressively train GameHunter's reasoning and execution in three stages, as shown in \cref{fig:pipeline9}.
Firstly, we establish reasoning and action generation capabilities.
We then train action execution conditioned on teacher-provided reasoning.
Finally, we jointly distill reasoning and actions on the student's autonomous trajectories.

\subsection{Progressive Training Strategy}
\label{Progressive Training Strategy}
\myPara{Progressive Grouping.}
To progressively build game capabilities from basic operations to long-horizon decision-making, we divide 42 games equally into seven groups according to operation types, reasoning requirements, and memory span.
This arrangement exposes execution learning to different operations and contexts and gradually extends it to longer chains of clues.

The combination of games within each group determines the training content at the current stage.
Let $p(c \mid g)$ denote the proportion of tasks in game $g$ that require capability type $c$.
When the six games participate equally, the within-group capability distribution is:

\begin{equation}
p_k(c) = \frac{1}{6}\sum_{g\in \mathcal{G}_k} p(c \mid g),
\end{equation}
where $\mathcal{G}_k$ denotes the $k$-th game group.
We gradually introduce tasks requiring more operation steps and longer chains of clues to expand the training curriculum~\cite{bengio2009curriculum,narvekar2020curriculum}.

Each training batch covers the game groups in sequence.
Let $\mathcal{S}_b$ denote the checkpoint stage range for the $b$-th batch.
The training order is defined as follows:

\begin{equation}
\mathcal{S}_b = \big[s_b^{\mathrm{start}},\, s_b^{\mathrm{end}}\big], \mathcal{C}_b : (\mathcal{G}_1, \mathcal{S}_b) \to \dots \to (\mathcal{G}_7, \mathcal{S}_b),
\end{equation}
where $s_b^{\mathrm{start}}$ and $s_b^{\mathrm{end}}$ denote the starting and ending checkpoints of the current batch, respectively.
We switch to the next group after completing the stage objective of the current group and begin the next batch after all groups are completed.
After parameter updates, we retain the state and history of each game so that newly learned execution capabilities can continue to be used in the existing interaction context.
The grouping and ordering of games are described in detail in the supplementary material.

\myPara{Cold Start.}
Building on progressive grouping, we adopt joint student--teacher training.
The student learns general reasoning and action generation across games and stages from the 42games of Game408 training data.
The teachers are multiple lightweight models trained for local story stages.
During subsequent stage-wise decoupling training, the student continues learning with the full game state, while the teacher changes with the checkpoint to provide local guidance matched to the current stage.
Before these later training stages, we first perform cold-start training for both the student and the teachers.
The 133 teacher models first undergo supervised fine-tuning (SFT) on the full Game408 dataset and are then further fine-tuned for three epochs on their respective checkpoint-stage data.
The student then undergoes four SFT epochs on the dataset of 42 games.

\subsection{Stage-wise Decoupling}

\myPara{Action Candidates and Evaluation.}
Action Bootstrapping uses stage-wise decoupling to condition on the current intent and focus training on its execution.
A teacher model trained for the current game stage provides reasoning $z_t$ and a verified reference action $a_t^T$ at state $x_t$.
With the teacher action masked out, the student independently samples four actions conditioned on $u_t=(x_t,z_t)$.
They form a five-candidate group together with the reference action.
The student need not regenerate reasoning at this stage.
Candidate feedback compares implementations of the same intent:

\begin{equation}
\begin{aligned}
\mathcal G_t &= \bigl(a_t^T,a_t^{(1)},\dots,a_t^{(4)}\bigr),\\
a_t^{(j)} &\overset{\mathrm{i.i.d.}}{\sim} \pi_{\theta_b}(\cdot\mid u_t), \quad j=1,\dots,4,
\end{aligned}
\end{equation}
where $\theta_b$ denotes the student parameters held fixed during sampling.
The teacher is switched as the game stage changes, and teacher actions advance the environment.
The student thus practices execution under the same decision conditions.
An intent can be fulfilled by multiple valid operations, so we evaluate candidates by action function and key parameters.
Let $\ell_t(a)\in\{0,1\}$ indicate call validity and functional compatibility.
Let $s_{t,k}(a)\in[0,1]$ be the soft match score for key parameter $k$.
The reward is:

\begin{equation}
R_t(a)=\ell_t(a)\min_{k\in\mathcal K_t}s_{t,k}(a).
\end{equation}
where $\mathcal K_t$ contains the required targets, content items, and control parameters.
Taking the minimum emphasizes the weakest-matching key parameter and prevents high scores on other attributes from compensating for it.
Correctness is checked separately by $c_t(a)\in\{0,1\}$.
It determines whether all operational requirements are met and does not require a soft reward of 1.
The teacher reference receives the full score, while student scores retain differences between implementations.

\begin{figure*}[t]
  \centering
  \setlength{\abovecaptionskip}{2pt}  \includegraphics[width=0.95\textwidth]{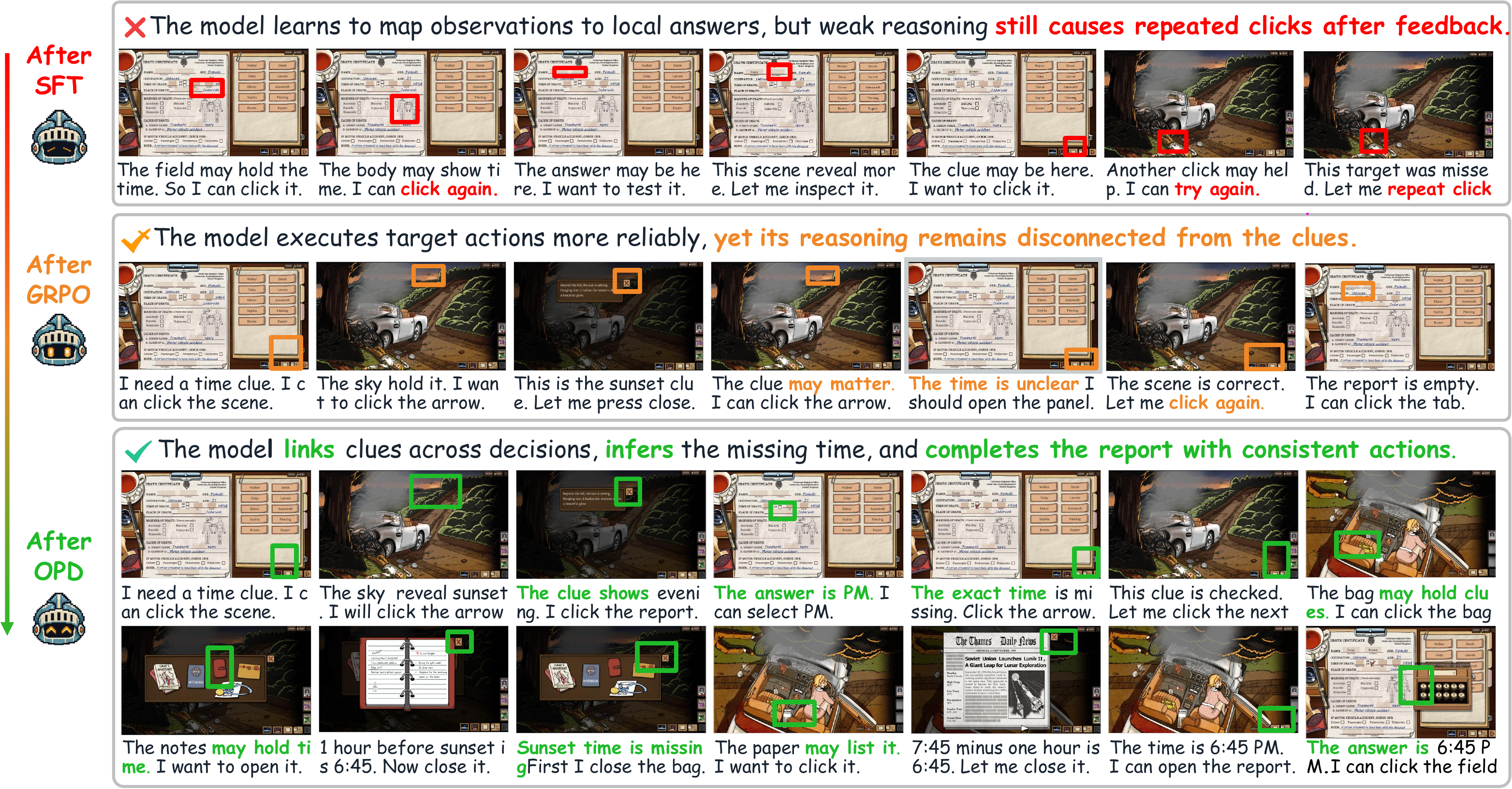}
\caption{\textbf{Qualitative comparison across three training stages.}
On the same investigation, SFT repeats clicks.
GRPO improves actions but leaves clues disconnected.
OPD links evidence across decisions to infer the missing time and complete the report.}

  \label{fig:COT22} %
\end{figure*}

\myPara{Advantage Truncation Constraint.}
After fixing the intent, group preferences must also respect operational correctness.
The partial matching score of an incorrect action may exceed the group mean.
It can therefore receive a positive relative advantage even when the group contains a correct teacher reference.
We first compute the mean $\mu_t$ and standard deviation $\sigma_t$ of the five candidate rewards.
We then normalize the student action rewards:
\begin{equation}
\widehat A_{t,j}=\frac{R_t(a_t^{(j)})-\mu_t}{\sigma_t+\delta}.
\end{equation}
where $\delta>0$ stabilizes normalization when candidate rewards are identical.
We constrain student advantages to prevent positive training weights for incorrect actions:

\begin{equation}
A_{t,j}=\widehat A_{t,j}-(1-c_{t,j})[\widehat A_{t,j}]_+,
\end{equation}
where $c_{t,j}=c_t(a_t^{(j)})$ and $[v]_+=\max(v,0)$.
Incorrect actions receive $A_{t,j}=\min(\widehat A_{t,j},0)$.
Correct actions retain their original relative advantages, including negative advantages when their rewards are below the group mean.
We do not recenter truncated advantages, as a mean shift may again give incorrect actions positive weights.

\myPara{GRPO Optimization.}
We combine cross-entropy supervision on the teacher action with correctness-constrained GRPO~\cite{shao2024deepseekmath}, concentrating direct output supervision on student actions.
Teacher reasoning is the input condition, and the teacher reference action is a valid target.
GRPO compares student implementations of the same intent:

\begin{equation}
\mathcal L_{\mathrm{action}}(\theta)=\mathcal L_{\mathrm{GRPO}}(\theta)-\lambda\log\pi_\theta(a_t^T\mid u_t).
\end{equation}
where $\lambda$ controls the weight of supervision from the teacher target.
GRPO adopts the clipped form of PPO~\cite{schulman2017proximal}.
The policy loss is computed only on the four student actions using the constrained advantages and likelihood ratios relative to $\pi_{\theta_b}$.
Stage-wise decoupling thus concentrates direct supervision on action outputs.
This builds an execution foundation for subsequent joint learning of reasoning and actions on autonomous trajectories.

\subsection{Autonomous Trajectories}
\myPara{LLM as a Judge.}
Full Bootstrapping starts from autonomous student trajectories: the teacher covers limited story paths, while the student may collect evidence in different orders.
GPT-5.6-Sol scores reasoning support, action selection, and execution consistency to guide intervention~\cite{zheng2023judge,liu2023geval,wang2023bias}.

\begin{equation}
Q_i = \sum_{d\in\mathcal{D}} q_i^{(d)}, \quad s_i = \mathcal{N}(Q_i) \in [0,1],
\end{equation}
where $q_i^{(d)}$ denotes the score for each criterion, and $\mathcal N$ is the normalization function.
A higher $s_i$ indicates better decision quality.
Human checks for GPT scoring and detailed rubrics appear in the supplementary material.

\begin{table*}[t]
  \small
  \setlength{\abovecaptionskip}{2pt}
  \tablestyle{3.45pt}{1.0}
  \newcommand{\gameTableBrand}[1]{%
  \raisebox{-.18em}{\includegraphics[width=1.22em,height=1.22em,keepaspectratio]{figsw/game408-icon-#1.png}}%
  \hspace{.40em}%
  }
  \newcommand{\gameTableBrandRoot}[1]{%
  \raisebox{-.18em}{\includegraphics[width=1.22em,height=1.22em,keepaspectratio]{game408-icon-#1.png}}%
  \hspace{.40em}%
  }
  \centering
  \caption{\textbf{Model comparison on Game408.}
Score measures task completion. Efficiency is the ratio of effective to total action steps.
Time reports average end-to-end action latency.
Action performance assesses action selection and target grounding.
Reasoning quality assesses logical validity, trajectory consistency, clue use, memory recall, and action assessment.
\texttt{--} denotes  unavailable values.}
  \begin{tabular}{lccccccccccc}
    \hline

    \hline

    \hline
    \multirow{2}{*}{\textbf{Models}}
    & \multirow{2}{*}{\textbf{Size}}
    & \multirow{2}{*}{\textbf{Score}}
    & \multirow{2}{*}{\textbf{Efficiency}}
    & \multirow{2}{*}{\makecell{\textbf{Time(s)}\,$\downarrow$}}
    & \multicolumn{2}{c}{\textbf{Action Performance (\%)}}
    & \multicolumn{5}{c}{\textbf{Reasoning Quality (\%)}} \\
    \cmidrule(l{2pt}r{2pt}){6-7}
    \cmidrule(l{2pt}r{2pt}){8-12}
    & & & & & \makecell{Action\\Selection} & \makecell{Target\\Grounding} & \makecell{Logic\\Valid} & \makecell{Traj.\\Consist} & \makecell{Clue\\Use} & \makecell{Mem.\\Recall} & \makecell{Action\\Judge} \\
    \midrule
    \gameTableBrand{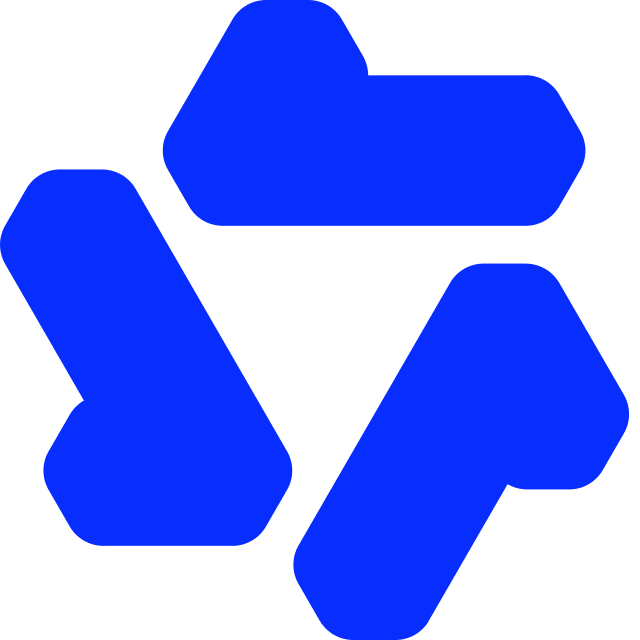}Qwen2.5-VL~\cite{bai2025qwen25vl} & 7B & 2.49 & 7.98 & 7.47 & 19.82 & 62.47 & 6.83 & 4.67 & 3.84 & 5.29 & 7.16 \\
    \gameTableBrand{qwen}Qwen3-VL-Instruct~\cite{bai2025qwen3vl} & 32B & 19.07 & 16.11 & 12.11 & 32.68 & 79.63 & 19.42 & 16.87 & 15.63 & 18.74 & 21.36 \\
    \gameTableBrand{qwen}Qwen3.5~\cite{qwen2026qwen35_9b} & 9B & 14.62 & 13.76 & 14.11 & 25.68 & 69.65 & 9.45 & 6.49 & 7.71 & 13.43 & 12.62 \\
    \gameTableBrand{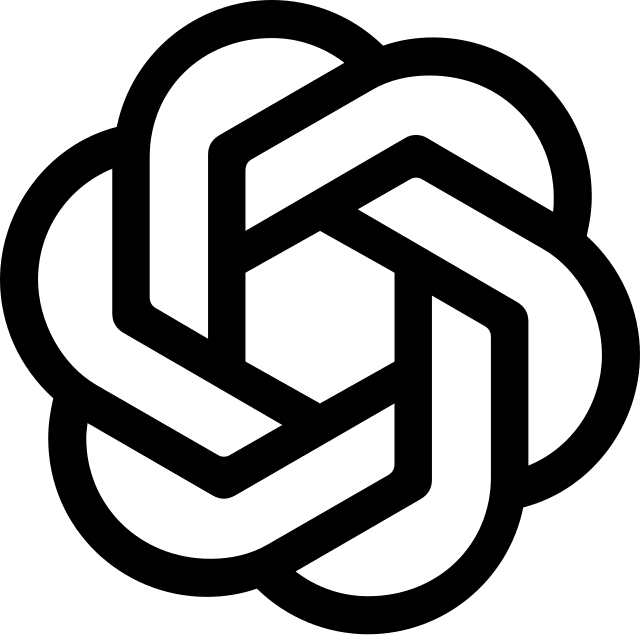}GPT-4o~\cite{openai2024gpt4o} & -- & 31.08 & 24.27 & 14.27 & 62.74 & 86.16 & 45.26 & 25.87 & 32.46 & 35.74 & 27.63 \\
    \gameTableBrand{openai}GPT-5.6-Terra~\cite{openai2026gpt56} & -- & \underline{53.48} & 45.57 & 14.53 & 82.13 & 90.42 & 59.73 & 46.82 & 53.16 & 57.34 & 48.67 \\
    \gameTableBrand{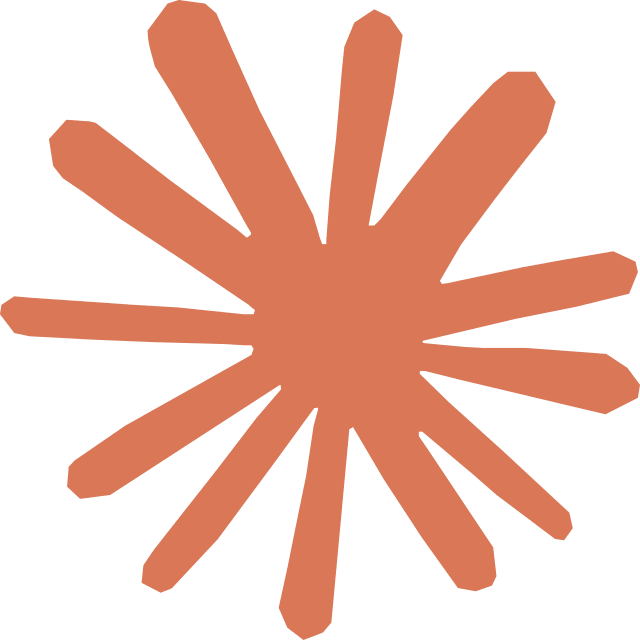}Claude-Haiku-4.5~\cite{anthropic2025haiku45} & -- & 37.78 & 62.55 & 9.29 & 71.84 & 87.12 & 53.46 & 31.84 & 36.72 & 40.63 & 33.27 \\
    \gameTableBrand{claude}Claude-Sonnet-4.6~\cite{anthropic2026sonnet46} & -- & 52.09 & \underline{76.58} & 15.42 & \underline{85.36} & \underline{92.08} & \underline{64.37} & \underline{49.18} & \underline{53.72} & 57.86 & \underline{60.74} \\
    \gameTableBrand{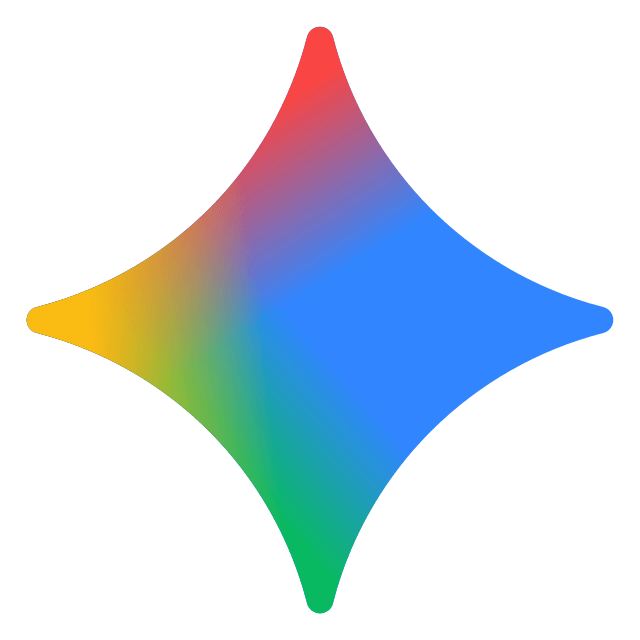}Gemini-3.7-Flash~\cite{doshi2026gemini37} & -- & 39.16 & 65.55 & 9.94 & 73.86 & 89.73 & 60.84 & 31.67 & 37.84 & \underline{58.73} & 33.46 \\
    \midrule
    \gameTableBrand{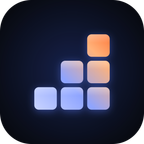}GameScaling-S3~\cite{zhang2026scaling} & 9B & 6.85 & 18.63 & \underline{4.37} & 38.76 & -- & -- & -- & -- & -- & -- \\
    \gameTableBrand{gamescaling}GameScaling-S3~\cite{zhang2026scaling} & 27B & 14.55 & 29.10 & 5.88 & 45.82 & -- & -- & -- & -- & -- & -- \\
    \gameTableBrand{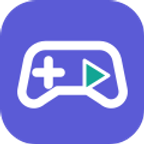}Game2Player~\cite{game2player2026} & 9B & 8.27 & 7.29 & 21.40 & 42.38 & 76.91 & 13.48 & 10.62 & 8.37 & -- & 11.94 \\
    \gameTableBrand{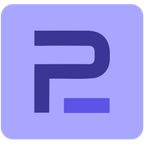}Open Pixel2Play~\cite{yue2026scaling} & 1.2B & 2.07 & 27.93 & \textbf{3.35} & -- & -- & -- & -- & -- & -- & -- \\
    \gameTableBrandRoot{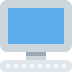}UI-TARS-1.5~\cite{bytedanceseed2025uitars15} & 7B & 10.58 & 25.96 & 11.07 & 48.73 & 78.46 & 15.83 & 12.46 & 9.72 & -- & 13.28 \\
    \gameTableBrandRoot{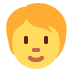}Human Players & -- & \textcolor{gray!65}{64.07} & \textcolor{gray!65}{84.72} & \textcolor{gray!65}{26.45} & \textcolor{gray!65}{97.55} & \textcolor{gray!65}{98.00} & \textcolor{gray!65}{81.46} & \textcolor{gray!65}{76.39} & \textcolor{gray!65}{69.00} & \textcolor{gray!65}{81.00} & \textcolor{gray!65}{93.46} \\
    \hdashline
    \rowcolor[HTML]{fff5f4}
    \gameTableBrand{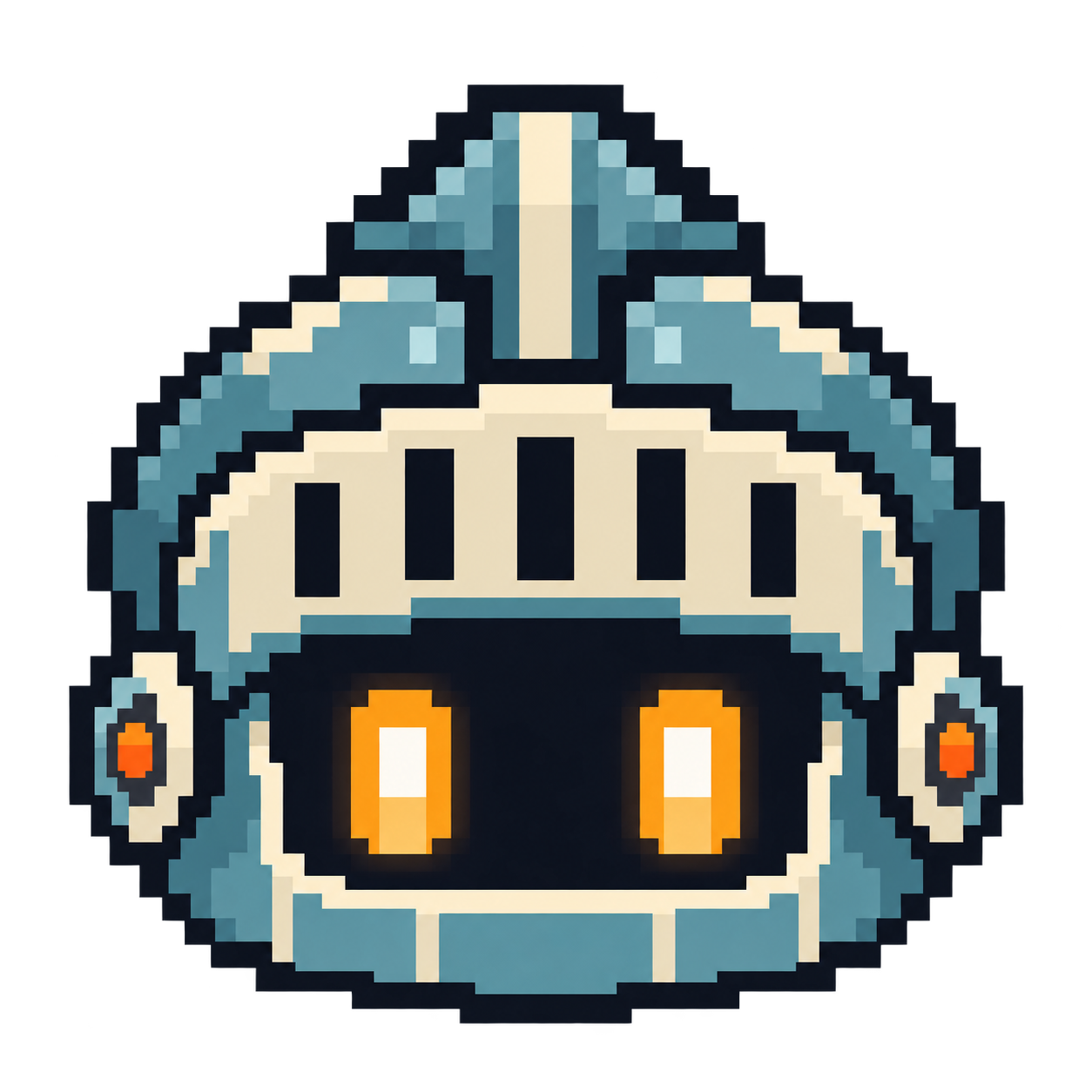}GameHunter-GRPO & 9B & 32.01 & 39.86 & 12.27 & 74.79 & 85.73 & 29.62 & 32.52 & 33.65 & 29.57 & 34.49 \\
    \rowcolor[HTML]{fff5f4}
    \gameTableBrand{robot}\textbf{GameHunter (Ours)} & 9B & \textbf{60.93} & \textbf{80.49} & 14.68 & \textbf{92.37} & \textbf{94.18} & \textbf{69.42} & \textbf{55.27} & \textbf{69.38} & \textbf{70.14} & \textbf{64.26} \\
    \hline

    \hline

    \hline
  \end{tabular}%

  \label{tab:game408_main_comparison}
\end{table*}

\myPara{Sliding Window Scoring.}
Single-step scores are sensitive to incidental actions, while full-trajectory evaluation retains the influence of earlier behavior.
We therefore use the product of scores over the latest 16 steps to measure recent quality, allowing reasonable exploration while triggering intervention after consecutive low-quality decisions.
Let $n_t=\min(16,t)$ and define:

\begin{equation}
\tau^\star = \min\Bigl\{\tau_{\mathrm{end}}, \inf\Bigl\{t \ge 1:\prod_{i=t-n_t+1}^{t} s_i < 0.32\Bigr\}\Bigr\},
\end{equation}
where $\tau_{\mathrm{end}}$ denotes the time when the task terminates, or the budget is exhausted.
Correction begins when the product falls below 0.32.
Earlier steps are left as the window advances.

\myPara{Local Distillation on Autonomous Trajectories.}
Training enters local distillation when the window identifies a recent decline in quality.
The training window comprises the current rollout's last 16 steps, or all available steps if fewer than 16.
Let $b^\star=\max\{1,\tau^\star-15\}$ and define:

\begin{equation}
\mathcal{W}^\star = \big((x_i, y_i)\big)_{i=b^\star}^{\tau^\star}, \quad q_{i,\ell}^T = q_{\phi_{\kappa_i}}(\cdot \mid x_i^T, y_{i,<\ell}),
\end{equation}
where $y_i$ contains reasoning and actions.
The teacher switches with the game stage $\kappa_i$, takes $x_i^T$ as input, and supervises both correct and erroneous steps conditioned on student-generated prefixes~\cite{agarwal2024gkd}.
Valid windows from different games jointly contribute to training:

\begin{equation}
\mathcal{L}_{\mathrm{full}} = \frac{1}{K}\sum_{g=1}^{K}\frac{1}{|\mathcal{I}_g|}\sum_{i\in\mathcal{I}_g}\frac{1}{|y_i|}\sum_{\ell=1}^{|y_i|} D_{i,\ell},
\end{equation}
where $K$ is the number of valid games, $\mathcal I_g$ contains decisions with reliable teacher supervision in game $g$'s window, and $D_{i,\ell}$ is the grouped forward KL loss for the corresponding token.
This objective jointly supervises reasoning and actions on the student's actual prefixes, allowing collected evidence to support subsequent intentions and operations.
After updates, we retain game state and history so that autonomous exploration resumes from the current state.

\myPara{Discussion.}
As shown in \cref{fig:32figures}, the student's free-exploration steps gradually increase during sliding-window OPD training.
This indicates that the mechanism can dynamically schedule teacher intervention based on the current trajectory state and support a gradual expansion of autonomous exploration.
We compare five OPD trigger thresholds.
Free-exploration steps increase under all five settings.
The 0.32 threshold reaches the highest level late in training and achieves the best Score, Reasoning, and Action performance.
We therefore adopt 0.32 for subsequent training on autonomous trajectories.
Local correction preserves the game state and interaction history, allowing exploration to resume without discarding previously acquired evidence.

\begin{table*}[t]
  \small
  \setlength{\abovecaptionskip}{2pt}
  \tablestyle{8.5pt}{1.0}
  \centering
  \caption{\textbf{Training ablations on Game408.}
Checkmarks indicate enabled training stages.
\bcancel{\checkmark} masks both teacher reasoning and actions.
Full Distill. trains after each 16-step rollout. Stepwise Distill. trains after each step.
Score and Efficiency follow \cref{tab:game408_main_comparison}.
Action is the mean of two action metrics. Reasoning is the mean of five reasoning metrics.
Avg.Time reports average end-to-end action latency in seconds.}
  \begin{tabular}{lcccccccc}
    \hline

    \hline

    \hline
    \multirow{2}{*}{\textbf{Models}}
    & \multicolumn{3}{c}{\textbf{Train}}
    & \multicolumn{5}{c}{\textbf{Metrics}} \\
    \cmidrule(l{2pt}r{2pt}){2-4}
    \cmidrule(l{2pt}r{2pt}){5-9}
    & \makecell{SFT}
    & \makecell{GRPO}
    & \makecell{OPD}
    & \makecell{Score}
    & \makecell{Efficiency}
    & \makecell{Avg.Time(s)$\downarrow$}
    & \makecell{Action}
    & \makecell{Reasoning} \\
    \midrule
    GameHunter w/o CoT & \checkmark & -- & -- & 16.21 & 34.76 & 20.18 & 49.26 & 16.36 \\
    GameHunter - SFT only & \checkmark & -- & -- & 18.74 & 38.23 & 24.46 & 53.47 & 27.74 \\
    GameHunter w/o Warm-up & \checkmark & -- & \checkmark & 42.63 & 54.27 & 18.53 & 78.46 & 42.36 \\
    GameHunter w/o Reasoning & \checkmark & \bcancel{\checkmark} & \checkmark & 54.18 & 63.42 & 17.26 & 82.24 & 52.72 \\
    \midrule
    GameHunter w/o Distill & \checkmark & \checkmark & -- & 32.01 & 39.86 & \textbf{12.27} & 80.26 & 31.97 \\
    GameHunter + Full Distill. & \checkmark & \checkmark & \checkmark & \underline{58.47} & \underline{74.83} & 17.18 & \underline{89.72} & \underline{61.26} \\
    GameHunter + Stepwise Distill. & \checkmark & \checkmark & \checkmark & 55.62 & 69.17 & 18.68 & 88.94 & 60.42 \\
    \hdashline
    \rowcolor[HTML]{fff5f4}
    \textbf{GameHunter (Ours)} & \checkmark & \checkmark & \checkmark & \textbf{60.93} & \textbf{80.49} & \underline{14.68} & \textbf{93.28} & \textbf{65.70} \\
    \hline

    \hline

    \hline
  \end{tabular}%

  \label{tab:game408_training_ablation}
\end{table*}

\begin{table}[t]
  \small
  \centering
  \setlength{\abovecaptionskip}{2pt}
  \tablestyle{4.5pt}{1.0}
  \caption{\textbf{Ablation of teacher size and supervision scope.}
Score measures task completion. Action and Reasoning follow \cref{tab:game408_training_ablation}.
Each checkpoint-stage teacher covers 3 checkpoints on average.}
  \begin{tabular}{llccc}
    \hline

    \hline

    \hline
    \multirow{2}{*}{\textbf{Teacher model}}
    & \multirow{2}{*}{\textbf{Scope}}
    & \multicolumn{3}{c}{\textbf{Student metrics (9B)}} \\
       \cmidrule(l{2.0pt}r{2pt}){3-5}
       & & \makecell{Score}
       & \makecell{Action}
       & \makecell{Reasoning} \\
    \midrule
    Qwen3.5-4B & All games & 16.17 & 33.47 & 10.71 \\
    Qwen3.5-9B & All games & 17.68 & 36.82 & 11.83 \\
    Qwen3.5-27B & All games & 19.23 & 39.64 & 13.46 \\
    Qwen3.5-4B & Per game & 46.63 & 74.18 & 32.38 \\
    Qwen3.5-9B & Per game & \underline{55.24} & \underline{85.86} & \underline{47.84} \\
    \hdashline
    \rowcolor[HTML]{fff5f4}
    \textbf{Qwen3.5-4B} & \textbf{checkpoints } & \textbf{60.93} & \textbf{93.28} & \textbf{65.70} \\
    \hline

    \hline

    \hline
  \end{tabular}%

  \label{tab:game408_teacher_ablation}
\end{table}

\section{Experiments}
\label{sec:game408_experiments}

\subsection{Experimental Details}

\myPara{Implementation Details.} GameHunter is fine-tuned from Qwen3.5-9B~\cite{team2026qwen35} and the teacher models are fine-tuned from Qwen3.5-4B.

\myPara{Evaluation Settings.} We evaluate on Game408-Val and score averages each game's completion rate over checkpoint-stage objectives and achievement goals.
Efficiency is the ratio of effective to total actions.
Time is the latency from model input to action completion.
Action selection and target grounding assess agreement with the current intent.
Reasoning quality covers logical validity, trajectory consistency, clue use, memory recall, and action assessment.
Details are in the supplementary material.
\subsection{Main Results}

\myPara{Target Alignment.}
As shown in \cref{tab:game408_main_comparison}, GameHunter achieves the highest Score, Action Selection, and Target Grounding among the evaluated models.
Compared with general-purpose models such as GPT-5.6-Terra and Claude-Sonnet-4.6, its advantage extends beyond visual recognition to translating target grounding into operations that match the intent.
This indicates that execution training conditioned on the teacher learns the correspondence among targets, control parameters, and intended outcomes, reducing ineffective operations despite correct target recognition.

\myPara{Sustained Reasoning.}
GameHunter-GRPO can already generate effective operations, but remains markedly weaker than the full model in clue use, memory recall, action assessment, and trajectory consistency.
These differences indicate that local trajectory distillation enables the student to reuse earlier evidence within its own interaction history.
The capabilities acquired through execution training can thus support subsequent investigation, rather than remaining limited to isolated actions.

\myPara{Human Evaluation.}
We also report evaluation metrics for human players as a reference for overall decision quality.
GameHunter is very close to human players on several metrics, but humans remain more accurate in action selection and target grounding and more coherent across the full trajectory.
These gaps mainly concern precise execution, evidence use across stages, and sustained recovery.

\subsection{Ablation Analysis}

\myPara{Component Ablation.}
We ablate the two training components in \cref{tab:game408_training_ablation}.
The results show that removing stage-wise decoupling training or distillation on autonomous trajectories markedly reduces the scores across metrics.
The \bcancel{\checkmark} variant also underperforms the full model, and the degradation is more pronounced with SFT alone.
These results indicate complementary roles for stage-wise decoupling and learning on autonomous trajectories.
The former develops execution capabilities under teacher-provided intent, while the latter jointly corrects reasoning and actions based on the student's own interaction history.
Focusing first on execution and then learning the coordination of reasoning and actions on autonomous trajectories helps translate individual operation capabilities into progress on long-horizon tasks.

\myPara{Necessity of the Sliding Window.}
In \cref{tab:game408_training_ablation}, we compare sliding window intervention based on recent trajectory quality with training at every step and every 16 steps to examine how intervention timing affects training.
The results show that the sliding window strategy outperforms both fixed intervention schedules in task completion and aggregate action and reasoning quality.
This is because the sliding window allows exploration to continue while recent trajectory quality remains at or above the threshold and triggers local distillation when it falls below the threshold, balancing room for exploration with timely correction.

\myPara{Discussion of Teacher Models.}
We fix the student model and vary teacher size and supervision scope across all games, individual games, and local checkpoint stages.
As shown in \cref{tab:game408_teacher_ablation}, narrowing the supervision scope consistently improves task completion when teacher size is fixed at 4B.
Checkpoint-stage teachers yield a higher student Score than a 27B all-game teacher and a 9B per-game teacher.
This indicates that supervision tailored to story stages is more effective than simply enlarging an all-game teacher.
Matching supervision to the student's current objectives, clues, and operations enables compact teachers to provide effective local guidance.

\subsection{Qualitative Comparison}

\begin{figure*}[!t]
  \centering
  \setlength{\abovecaptionskip}{2pt}
  \includegraphics[width=0.95\textwidth]{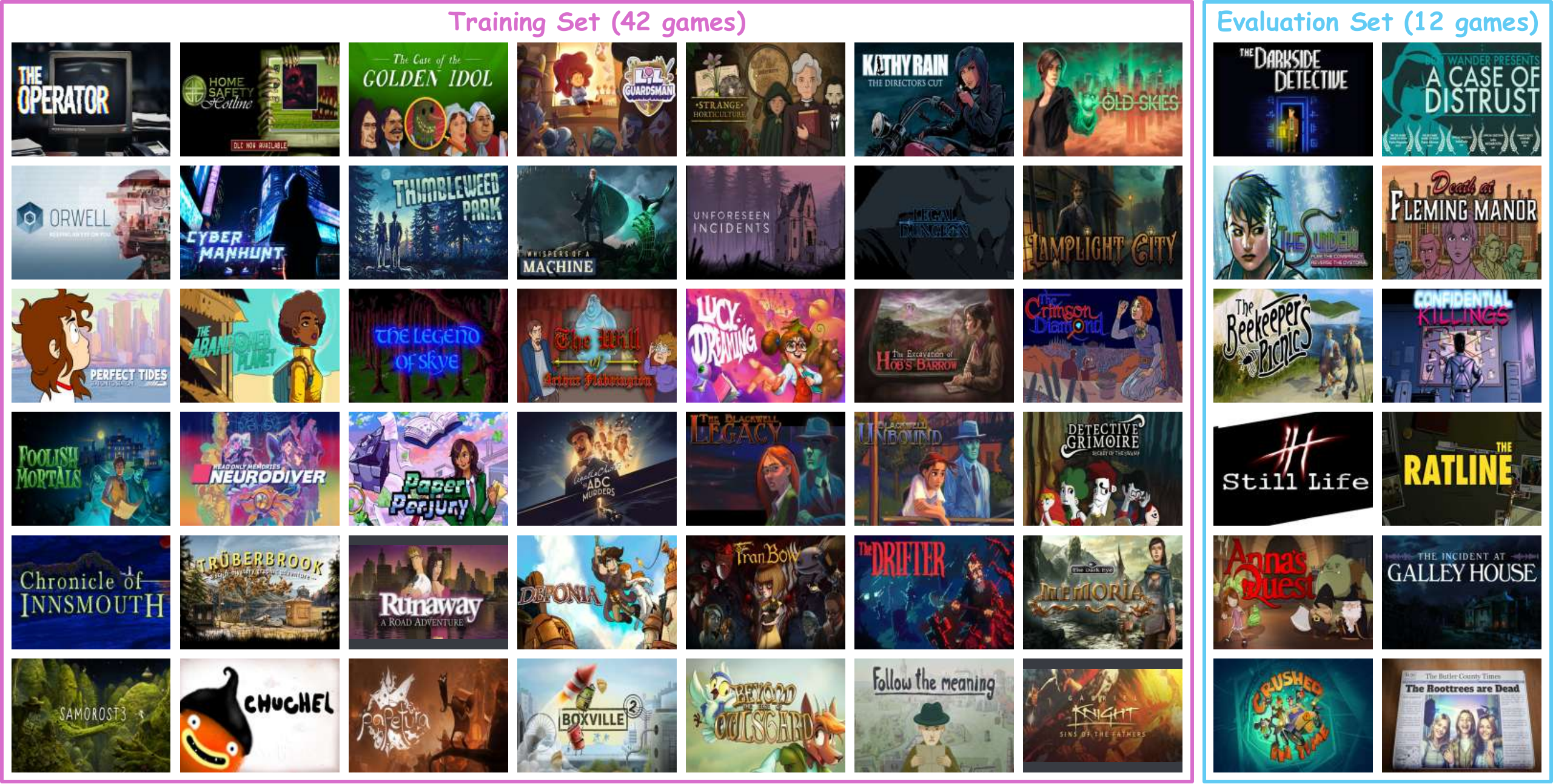}
  \caption{\textbf{Cover art for the 54 entries in the Game408 catalog.}}
  \label{fig:sup-game-catalog}
\end{figure*}

\cref{fig:COT22} compares reasoning and actions in the same game across three training stages.
After SFT, the model generates basic reasoning from observations and attempts actions but repeatedly lingers in place, indicating weak reasoning--action coordination.
Stage-wise decoupling through GRPO makes actions more reasonable.
Local distillation on autonomous trajectories then enables the model to correct deviations and maintain coherence in subsequent decisions.
Together, these observations support the effectiveness of GameHunter's staged training in extending local execution capabilities into sustained reasoning.

\section{Conclusions}

We introduce Game408 with 54 games, 408 checkpoints, and over 24K human reasoning and action records.
For long-horizon narrative games, GameHunter combines progressive training, student--teacher cold start, stage-wise decoupling, and local distillation on autonomous trajectories to connect reasoning and actions.
Experiments show improved action reliability and sustained reasoning.
Ablations support the complementary roles of stage-wise decoupling and local distillation on autonomous trajectories.
Together, these components turn reliable local actions into sustained progress across long-horizon tasks.

\section*{LLM Usage Statement}
We use ChatGPT to assist with generating code and related implementations.
The authors review, edit, or approve the resulting code and implementations.
These tools also assist with editing figures and visual materials created by the authors.
During dataset construction, we use GPT to assist with data generation and use related tools to edit or standardize data and annotations.
The authors check all content generated or assisted by these tools and take responsibility for the final manuscript.

\appendix
\section*{Appendix}
\counterwithin{table}{section}
\counterwithin{figure}{section}
\counterwithin{equation}{section}
\makeatletter
\setlength{\@fptop}{0pt}
\setlength{\@dblfptop}{0pt}
\setlength{\@fpsep}{10pt}
\setlength{\@dblfpsep}{10pt}
\makeatother

\section{Implementation Details}
\label{sec:sup-implementation}

We train on eight NVIDIA RTX PRO 6000 GPUs using \texttt{verl}~\cite{sheng2025hybridflow}, Fully Sharded Data Parallelism (FSDP)~\cite{zhao2023fsdp}, and AdamW~\cite{loshchilov2019adamw}.
We set the learning rate to $10^{-5}$ and use LoRA~\cite{hu2022lora} with rank 64 and alpha 128.
All evaluations use four NVIDIA A40 GPUs.

\section{Game Data}
\label{sec:sup-data}

\subsection{Game Categories}
\label{sec:sup-game-groups}

\begin{table*}[!t]
  \centering
  \setlength{\abovecaptionskip}{2pt}
  \caption{\textbf{Six representative task types and the skills they require.}}
  \label{tab:sup-game-types}
  \tablestyle{3.45pt}{1.0}
  \begin{tabular}{P{\dimexpr.22\linewidth-\tabcolsep\relax}P{\dimexpr.53\linewidth-4\tabcolsep\relax}P{\dimexpr.25\linewidth-\tabcolsep\relax}}
    \hline

    \hline

    \hline
    \textbf{Task type} & \textbf{Required skill} & \textbf{Example games} \\
    \specialrule{\lightrulewidth}{0pt}{0pt}
    \rowcolor{AppendixRowGray}
    Interface and document investigation
    & Locate relevant fields in records, databases, or dialogue transcripts, cross-check what they say about people or events, and submit evidence or act on it.
    & \textit{The Operator}, \textit{Legal Dungeon} \\
    \rowcolor{white}
    Investigation and case reconstruction
    & Compare testimony, scene observations, and physical evidence to infer relationships, chronology, and causality.
    & \textit{The Case of the Golden Idol} \\
    \rowcolor{AppendixRowGray}
    Character dialogue and choices
    & Use character relationships, earlier statements, and the current branch to decide when to follow up and which dialogue option can fill an information gap.
    & \textit{A Case of Distrust} \\
    \rowcolor{white}
    Item combination and contextual puzzles
    & Track where items came from and what they do, check inventory and prerequisites, and choose what to combine, exchange, or hand over.
    & \textit{The Legend of Skye} \\
    \rowcolor{AppendixRowGray}
    Scene exploration and mechanisms
    & Track locations and mechanism states across scenes and apply earlier clues to subsequent operations.
    & \textit{The Abandoned Planet}, \textit{Boxville 2} \\
    \rowcolor{white}
    Visual clues and sequential interaction
    & Infer order and correspondences from images or scene states to select the next interaction.
    & \textit{Read Only Memories}, \textit{CHUCHEL} \\
    \hline

    \hline

    \hline
  \end{tabular}
\end{table*}

We select 54 long-horizon reasoning games from Steam, each with a clear main storyline, and have human players verify the completion routes in actual gameplay.
The games primarily use two-dimensional interfaces, where players explore by inspecting screens and moving between scenes and can pause to review clues before acting.
We define six task types based on the evidence and interactions required by representative stage objectives as shown in \cref{tab:sup-game-types}).

\subsection{Game Grouping}

We form nine groups, each with one game from each of the six task types.
The first seven groups are used for training, and the last two are held out exclusively for testing.
In each training round, we visit the training groups in order at the current checkpoint stage so that the model practices the reasoning and interaction skills required by all six task types.
As the stories progress, later tasks require integrating longer chains of clues across scenes.
For segments that require rapid clicking or a sequence of actions, we use scripts to perform the required actions and advance past them. These segments are excluded from scoring.

\subsection{Game Catalog}

Following the description of game grouping, we present all 54 entries in the Game408 catalog in \cref{fig:sup-game-catalog}.
Gameplay recordings from this collection form the basis for the annotations described in the next section.

\subsection{Checkpoint Segmentation}
\label{sec:sup-collection}

We organize 34 volunteers by their gameplay experience.
Experienced volunteers map the main and side storylines and use Steam achievement progress to identify 408 checkpoints, taking into account the difficulty of adjacent stages and the span of the required evidence.
In \textit{The Operator}, checkpoints are placed around the transitions between workdays, advances in the evidence chain, and the events immediately before and after bomb defusal.
In \textit{The Case of the Golden Idol}, the boundaries mark the opening of a case, the recovery of clues, and case closure.
Novice volunteers playtest the stages to check that the objectives and starting information are clear and that the difficulty is appropriate.

\section{Data Collection}
\label{sec:sup-annotation}
\subsection{Gameplay Recordings }
\label{sec:sup-horizon}

\begin{table}[!t]
  \centering
  \setlength{\abovecaptionskip}{2pt}
  \caption{\textbf{Reasoning horizons and required evidence.}}
  \label{tab:sup-horizon}
  \tablestyle{4.5pt}{1.0}
  \begin{tabular}{P{\dimexpr.11\linewidth-\tabcolsep\relax}P{\dimexpr.89\linewidth-3\tabcolsep\relax}}
    \hline

    \hline

    \hline
    \textbf{Level} & \textbf{Evidence needed for the action} \\
    \specialrule{\lightrulewidth}{0pt}{0pt}
    \rowcolor{AppendixRowGray}
    D0 & The current observation suffices for a basic action. \\
    \rowcolor{white}
    D1 & The decision also requires a recent action or a clue from the same scene. \\
    \rowcolor{AppendixRowGray}
    D2 & The choice depends on multiple clues, item prerequisites, or earlier records within a local story segment. \\
    \rowcolor{white}
    D3 & The choice draws on evidence, branch states, or long-term goals across scenes or stages. \\
    \hline

    \hline

    \hline
  \end{tabular}
\end{table}

We collect more than 260 hours of human gameplay recordings across the 54 games and cross-check puzzle solutions against online videos, gameplay streams, and community walkthroughs.
GPT-5.6~\cite{openai2026gpt56} extracts candidate actions. Annotators verify each action and its frame in sequence.
Each Game408-CoT annotation covers a single semantic action and records the pre-action observation, relevant earlier clues, and the action intent, explaining how the evidence supports the operation.
We classify reasoning into four horizons, as shown in \cref{tab:sup-horizon}.
The adjacent records follow the actual order of play, yielding more than 24,000 human annotations.
After human review, GPT-4o~\cite{openai2024gpt4o} formats each annotation as reasoning and action.

\subsection{Exploration and recovery from failures}

\label{sec:sup-volunteer-data}
Game408-CoA retains actions that advance the story efficiently as well as genuine suboptimal attempts, such as opening a file at the console to see what it contains, trying a newly acquired tool, or selecting the wrong interaction target.
Volunteers construct examples of correction and verification that show how a model can recover after a failed attempt and where it can look for clues when evidence is insufficient.

\subsection{Reflection after repeated actions}

Our reflection examples cover both repeated memory queries without keyboard or mouse input and repeated keyboard or mouse actions.
One example illustrates a change of strategy with the following reflection.
\textquotedblleft The last three clicks on the magnifying glass left the screen unchanged, so I should try another approach, such as clicking the map.\textquotedblright

\section{Evaluation and Scoring}
\label{sec:sup-eval}

\begin{table*}[!t]
  \centering
  \setlength{\abovecaptionskip}{2pt}
  \caption{\textbf{Scoring criteria for actions and reasoning.}}
  \label{tab:sup-criteria}
  \tablestyle{3.45pt}{1.0}
  \begin{tabular}{P{\dimexpr.11\linewidth-\tabcolsep\relax}P{.19\linewidth}P{.37\linewidth}P{\dimexpr.33\linewidth-7\tabcolsep\relax}}
    \hline

    \hline

    \hline
    \textbf{Group} & \textbf{Criterion} & \textbf{Full-credit condition} & \textbf{Typical deductions} \\
    \specialrule{\lightrulewidth}{0pt}{0pt}
    \rowcolor{AppendixRowGray}
    Action & Action Selection
    & The chosen operation fits the subgoal and known constraints and can advance the task or distinguish hypotheses.
    & The choice overlooks a prerequisite or repeats a known ineffective attempt. \\
    \rowcolor{white}
    Action & Target Grounding
    & The call targets an object or an equivalent control that matches the intent.
    Check the object for a click, both endpoints for a drag, the input field for typing, and the pane for scrolling.
    & A wrong critical target scores zero, as when the right key is dragged to the wrong lock.
    Action selection is scored independently. \\
    \rowcolor{AppendixRowGray}
    Reasoning & Logical Validity
    & The conclusion follows from known facts and stated assumptions, with evidence for the key causal links.
    & The reasoning omits a key premise or draws a conclusion that the observations do not support. \\
    \rowcolor{white}
    Reasoning & Trajectory Consistency
    & The reasoning reflects the current task state, prior outcomes, and remaining prerequisites.
    & It uses an outdated state or skips a prerequisite step. \\
    \rowcolor{AppendixRowGray}
    Reasoning & Clue Use
    & The reasoning shows how clues support the choice or rule out alternatives.
    & It omits a decisive clue or repeats a clue without applying it to the choice. \\
    \rowcolor{white}
    Reasoning & Memory Recall
    & The reasoning accurately recalls an earlier fact that remains relevant and connects it to the decision.
    & It misremembers an earlier fact or omits a memory needed for the current step. \\
    \rowcolor{AppendixRowGray}
    Reasoning & Action Judgment
    & Pre-action evidence supports the stated prerequisites and expected effect, with unverified effects expressed as expectations.
    & It treats an interface affordance as proof that the expected effect will occur. \\
    \hline

    \hline

    \hline
  \end{tabular}
\end{table*}
\subsection{Model Evaluation Metrics}
\label{sec:sup-criteria}

On Game408-Val, we compute each game's completion rate over its checkpoint-stage objectives and achievement goals, then take the arithmetic mean of these rates across games to obtain the task completion score (\textbf{Score}).
\textbf{Efficiency} is the fraction of model actions that are effective, including necessary navigation, evidence collection, and tests supported by the available evidence.
\textbf{Time} is the mean latency from model input to action completion.
An LLM judge~\cite{zheng2023judge,liu2023geval} scores each decision on our two action and five reasoning criteria (\cref{tab:sup-criteria}). It checks the reasoning and semantic tool call against the frame, task state, history, and memories available before the action.

\subsection{Single-Step Scoring During Training}
\label{sec:sup-training-scoring}

\begin{table}[!t]
  \centering
  \setlength{\abovecaptionskip}{2pt}
  \caption{\textbf{Common score levels for the seven criteria.}}
  \label{tab:sup-scale}
  \tablestyle{4.5pt}{1.0}
  \begin{tabular}{P{\dimexpr.12\linewidth-\tabcolsep\relax}P{\dimexpr.88\linewidth-3\tabcolsep\relax}}
    \hline

    \hline

    \hline
    \textbf{Score} & \textbf{Criterion-level judgment} \\
    \specialrule{\lightrulewidth}{0pt}{0pt}
    \rowcolor{AppendixRowGray}
    5 & Meets the criterion, with sufficient evidence for the key judgment or operation. \\
    \rowcolor{white}
    4 & A minor error leaves the central judgment or operation intact. \\
    \rowcolor{AppendixRowGray}
    3 & The direction is broadly sound, but an important clue or condition is missing. \\
    \rowcolor{white}
    2 & A major gap leaves only part of the judgment or operation supported. \\
    \rowcolor{AppendixRowGray}
    1 & Only a small part is correct and the central judgment or operation is wrong. \\
    \rowcolor{white}
    0 & Contradicts explicit evidence or a task constraint, or omits information essential to the step. \\
    \hline

    \hline

    \hline
  \end{tabular}
\end{table}

We group the seven ratings in \cref{tab:sup-criteria} into three scores.
Reasoning support sums the raw scores for Logical Validity, Trajectory Consistency, Clue Use, and Memory Recall.
Action selection uses the raw score for the corresponding criterion.
Execution consistency sums the raw scores for Target Grounding and Action Judgment.
When all seven criteria are scored, the three maximum scores are 20, 5, and 10, respectively, yielding a single-step quality score of $s_i=Q_i/35$ for their sum $Q_i$.
At step 482 in \cref{sec:sup-scoring-cases}, the three scores are 18, 5, and 10, giving $s_i=33/35\approx 0.9429$.

\subsection{Sliding Window Trigger}
\label{sec:sup-window}

\begin{figure*}[!t]
  \centering
  \renewcommand{\thefigure}{\getrefnumber{sec:sup-review}.\arabic{figure}}
  \setlength{\abovecaptionskip}{2pt}
  \includegraphics[width=0.98\textwidth]{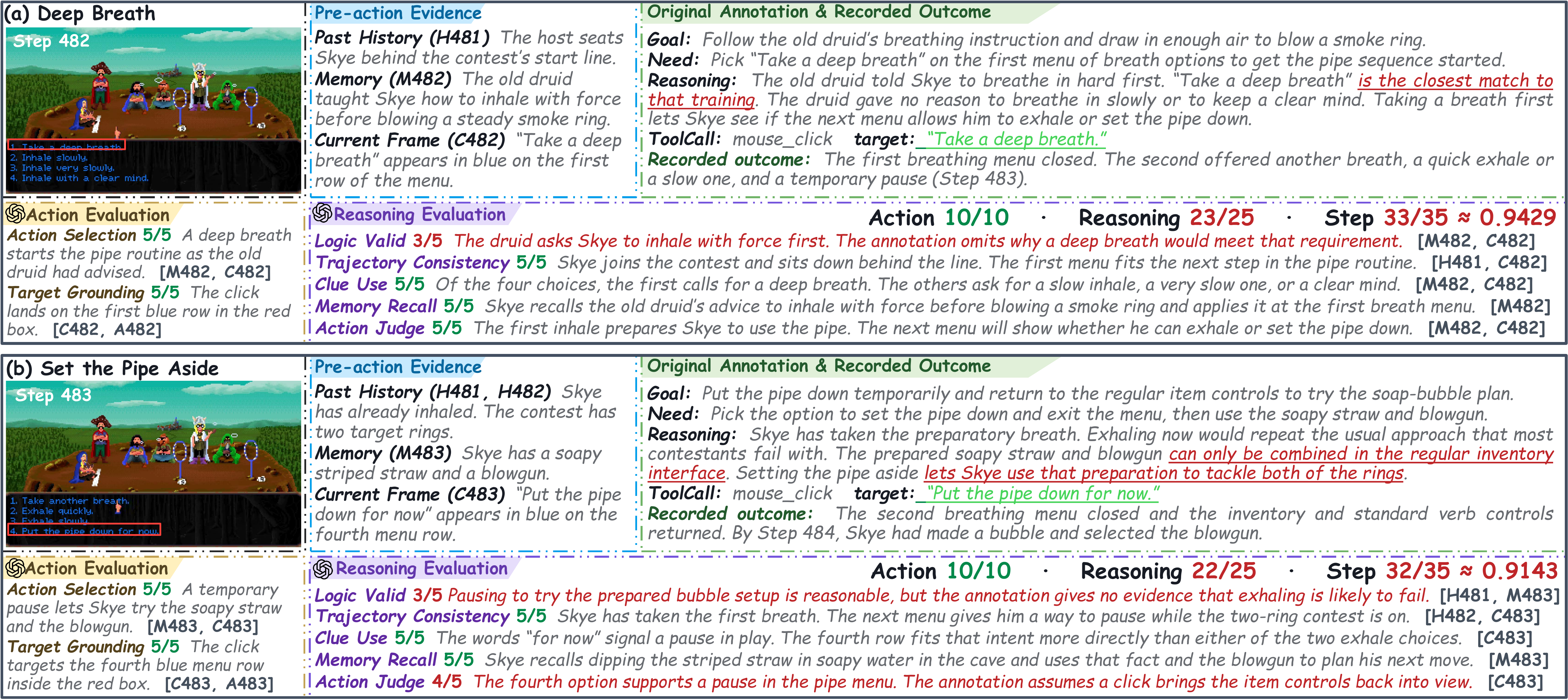}
  \caption{\textbf{Pre-action evidence, original annotations, recorded outcomes, and seven criterion scores for two consecutive steps in \textit{The Legend of Skye}.}
  H, M, C, and A denote earlier records, recalled memories, the current frame, and the tool call, respectively, and the numbers are the original action IDs.}
  \label{fig:sup-step-examples} %
\end{figure*}

\begin{figure}[!t]
  \centering
  \renewcommand{\thefigure}{\getrefnumber{sec:sup-review}.\arabic{figure}}
  \setlength{\abovecaptionskip}{2pt}
  \includegraphics[width=0.85\columnwidth]{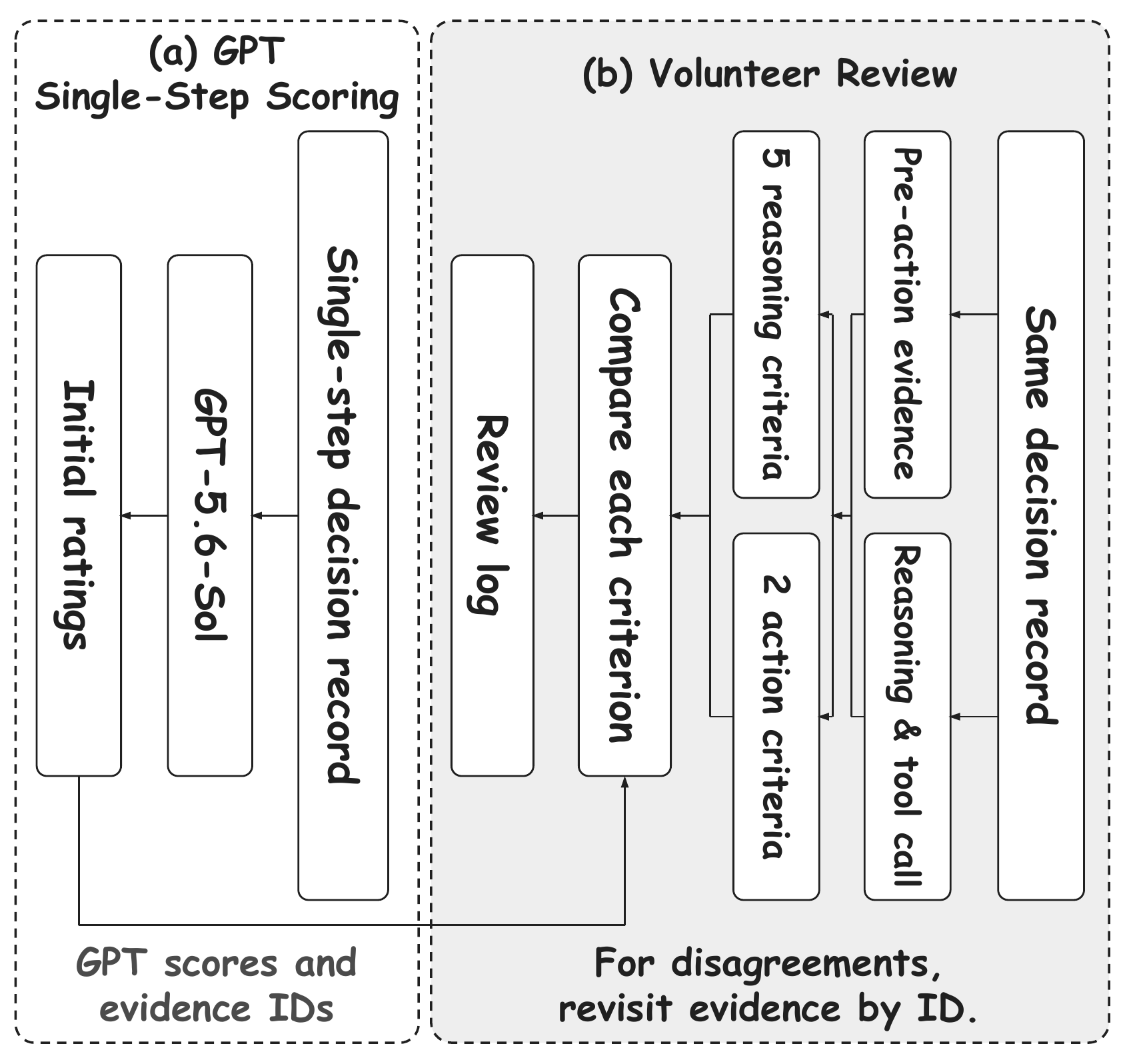}
  \caption{\textbf{GPT single-step scoring and volunteer review.}}
  \label{fig:sup-review} %
\end{figure}

We compute the product of the single-step quality scores over the most recent 16 steps.
If fewer than 16 steps are available, we use all available steps.
Teacher intervention is triggered the first time the product falls strictly below $0.32$.
Several low scores can compound and reduce the window product, and a sufficiently low score on a single step can also trigger intervention.
Each \texttt{memory\_search} call counts as one step in the window.

For example, if 15 steps in the window receive full credit and the remaining step scores $10/35$, the window product is $10/35 \approx 0.286$, below the trigger threshold.

\subsection{Single-Step Scores and Aggregation}
\label{sec:sup-aggregation}

For each applicable criterion it can assess, GPT-5.6-Sol~\cite{openai2026gpt56} assigns an integer from 0 to 5 and cites supporting evidence.
All seven criteria use the scale in \cref{tab:sup-scale}.

When all seven criteria apply and can be scored, we normalize the action score $A_i$, reasoning score $R_i$, and overall score $s_i$ for step $i$ by their respective maxima.
Each criterion has equal weight, giving group weights of $2/7$ for action and $5/7$ for reasoning.
\begin{equation}
  \begin{aligned}
    A_i &= \frac{q_i^{\mathrm{AS}}+q_i^{\mathrm{TG}}}{10},\\
    R_i &= \frac{q_i^{\mathrm{LV}}+q_i^{\mathrm{TC}}+q_i^{\mathrm{CU}}
      +q_i^{\mathrm{MR}}+q_i^{\mathrm{AJ}}}{25}.
  \end{aligned}
  \label{eq:sup-group-scores}
\end{equation}
\begin{equation}
  \begin{aligned}
    s_i &=
    \frac{q_i^{\mathrm{AS}}+q_i^{\mathrm{TG}}+q_i^{\mathrm{LV}}
      +q_i^{\mathrm{TC}}+q_i^{\mathrm{CU}}+q_i^{\mathrm{MR}}+q_i^{\mathrm{AJ}}}{35} \\
    &= \frac{2A_i+5R_i}{7}.
  \end{aligned}
  \label{eq:sup-step-score}
\end{equation}

We label a step excellent if its unrounded overall score is strictly greater than $0.93$.
With all seven criteria scored, this requires at least 33 raw points.

We report an aggregate percentage for each of the seven criteria.
Let $I_d$ be the set of steps in the evaluation set for which criterion $d$ applies and can be scored.
The percentage for criterion $d$ is computed as follows.
\begin{equation}
  P_d=\frac{100}{5\,|I_d|}\sum_{i\in I_d} q_i^{(d)}.
  \label{eq:sup-percent}
\end{equation}
\textbf{Action} and \textbf{Reasoning} are the arithmetic means of the two action percentages and the five reasoning percentages, respectively.
\textbf{Score} measures task completion as defined in \cref{sec:sup-criteria}.

\section{Real Scoring Examples}
\label{sec:sup-review}
\subsection{Volunteer Review}
\label{sec:sup-volunteer-review}

The volunteers and GPT-5.6-Sol use the same single-step decision record as shown in \cref{fig:sup-review}.
The volunteers first independently score the seven criteria in \cref{sec:sup-criteria}, then compare their ratings and cited evidence with the GPT judge's for each criterion.
For disagreements, they use evidence IDs to revisit the pre-action frame, earlier clues, or semantic tool call and record the affected criteria.
For each criterion, exact agreement is the number of matching score pairs before adjudication divided by the number of pairs for which both raters provide a valid score.
Overall agreement is the total number of matching pairs across the seven criteria divided by the total number of valid pairs.
We sampled 50\% of the trajectories for review, achieving 85\% overall agreement.

\subsection{Single-Step Scoring Examples}
\label{sec:sup-scoring-cases}

\cref{fig:sup-step-examples} shows two consecutive decisions in the smoke-ring contest: taking a deep breath at step 482 and setting the pipe aside at step 483.
Both steps are scored using only the evidence available before the action, yielding overall scores of $33/35$ and $32/35$, respectively.
The two steps receive the same action score, and the difference comes from Action Judgment because the pause option at step 483 does not establish that clicking it will restore the inventory controls.

\section{Continuous Decision Records}

We present 12 consecutive steps from each of two games, showing the pre-action frame, task state, reasoning, and tool call for each step.

\subsection{\textit{Orwell: Keeping an Eye On You}}
\label{sec:sup-orwell}

In steps 182–193, the investigator uses emails and chat logs to update Nina Maternova’s profile. An earlier relationship record then helps identify Desmond Schwartz in a military memorial.

\subsection{\textit{The Abandoned Planet}}
\label{sec:sup-planet}

In steps 345–356, the player matches cactus juice and a red flower to their corresponding basins. At the flame-marked basin, the player places a pine cone and then selects cloth as possible tinder.

\clearpage

\begin{figure*}[!t]
  \centering
  \setlength{\abovecaptionskip}{2pt}
  \includegraphics[width=\textwidth,height=\dimexpr\textheight-16pt\relax,keepaspectratio]{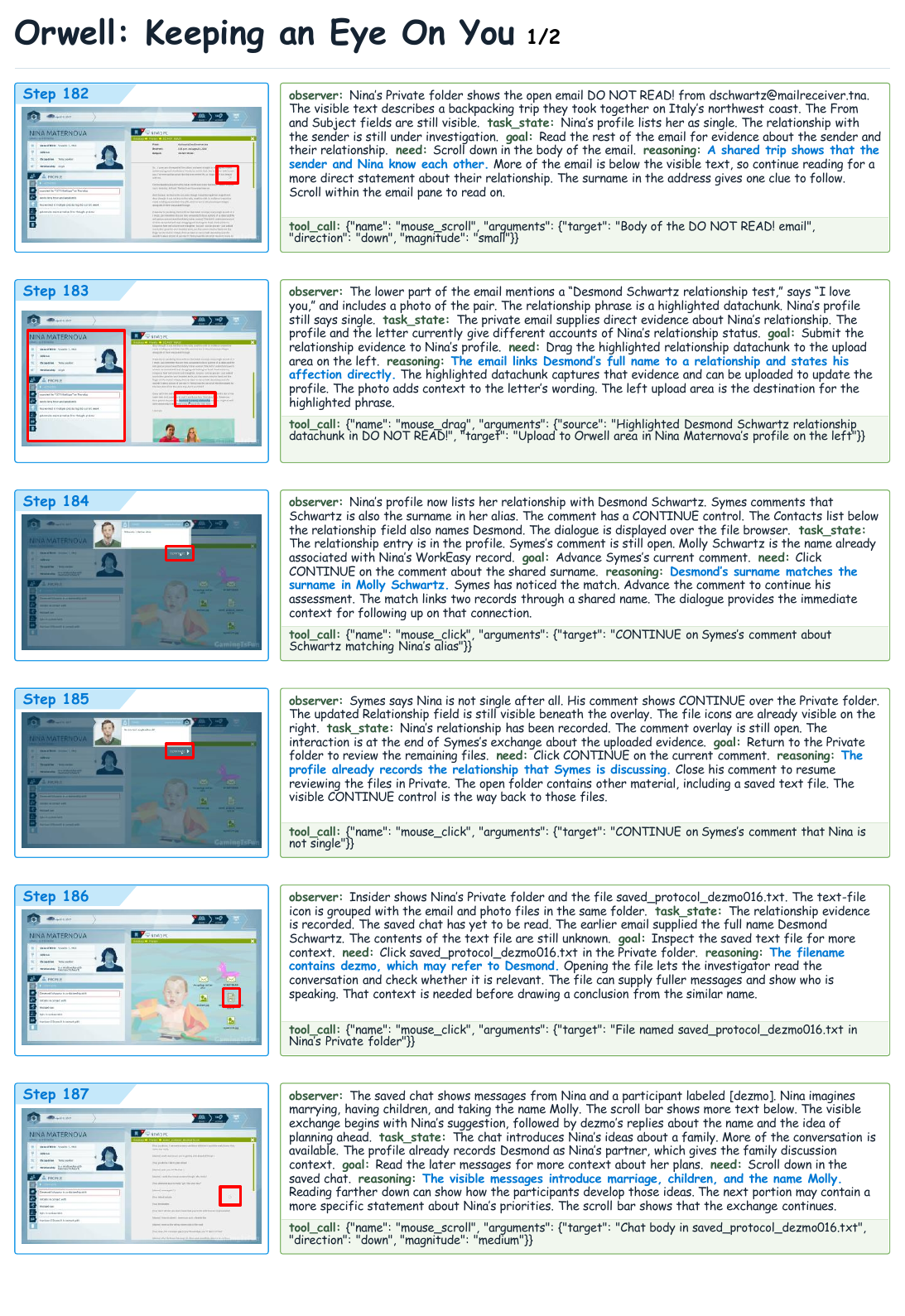}
  \caption{\textbf{\textit{Orwell: Keeping an Eye On You}, original steps 182--187.}}
  \label{fig:sup-orwell-one} %
\end{figure*}

\clearpage

\begin{figure*}[!t]
  \centering
  \setlength{\abovecaptionskip}{2pt}
  \includegraphics[width=\textwidth,height=\dimexpr\textheight-16pt\relax,keepaspectratio]{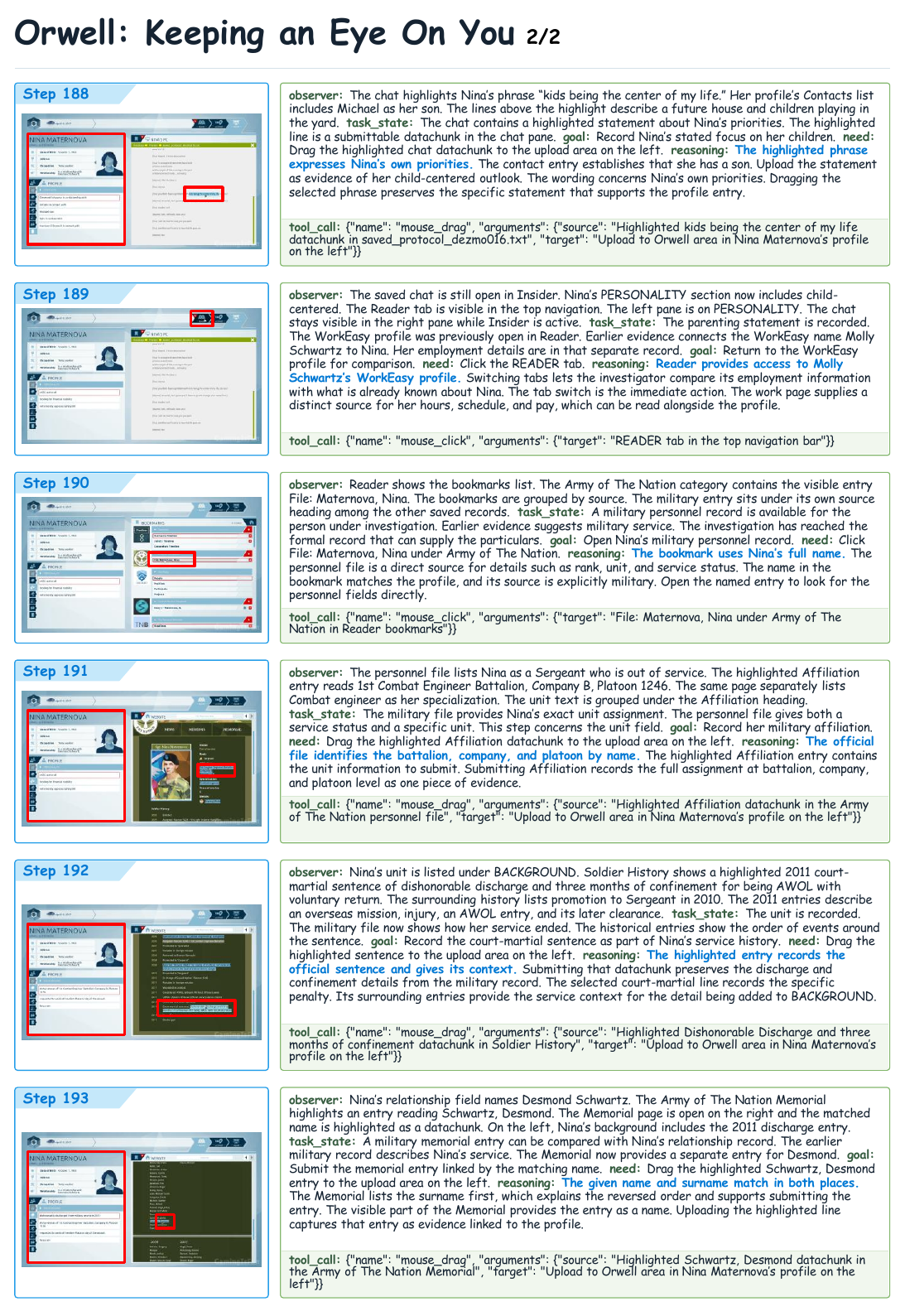}
  \caption{\textbf{\textit{Orwell: Keeping an Eye On You}, original steps 188--193.}}
  \label{fig:sup-orwell-two} %
\end{figure*}

\clearpage

\begin{figure*}[!t]
  \centering
  \setlength{\abovecaptionskip}{2pt}
  \includegraphics[width=\textwidth,height=\dimexpr\textheight-16pt\relax,keepaspectratio]{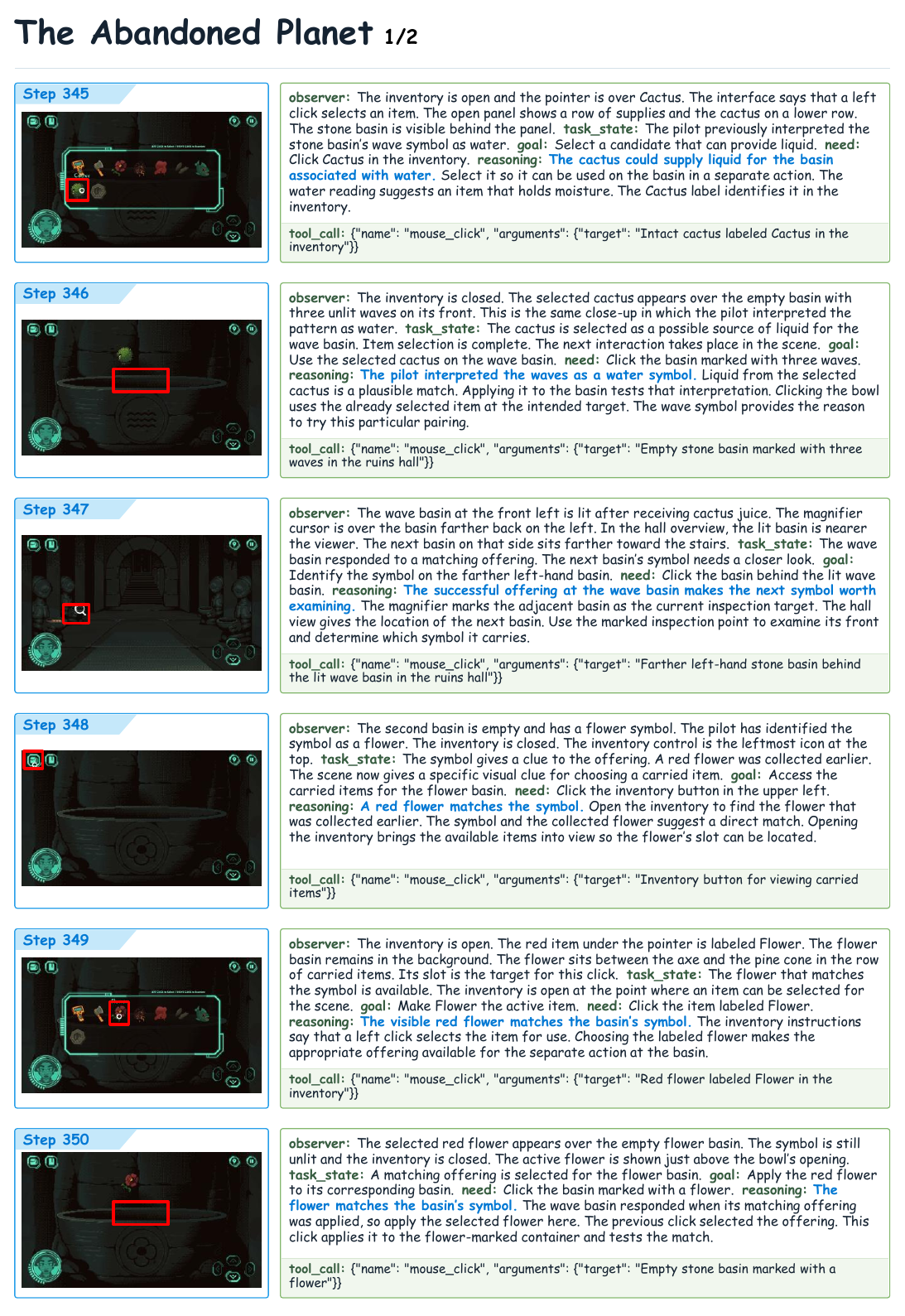}
  \caption{\textbf{\textit{The Abandoned Planet}, original steps 345--350.}}
  \label{fig:sup-planet-one} %
\end{figure*}

\clearpage

\begin{figure*}[!t]
  \centering
  \setlength{\abovecaptionskip}{2pt}
  \includegraphics[width=\textwidth,height=\dimexpr\textheight-16pt\relax,keepaspectratio]{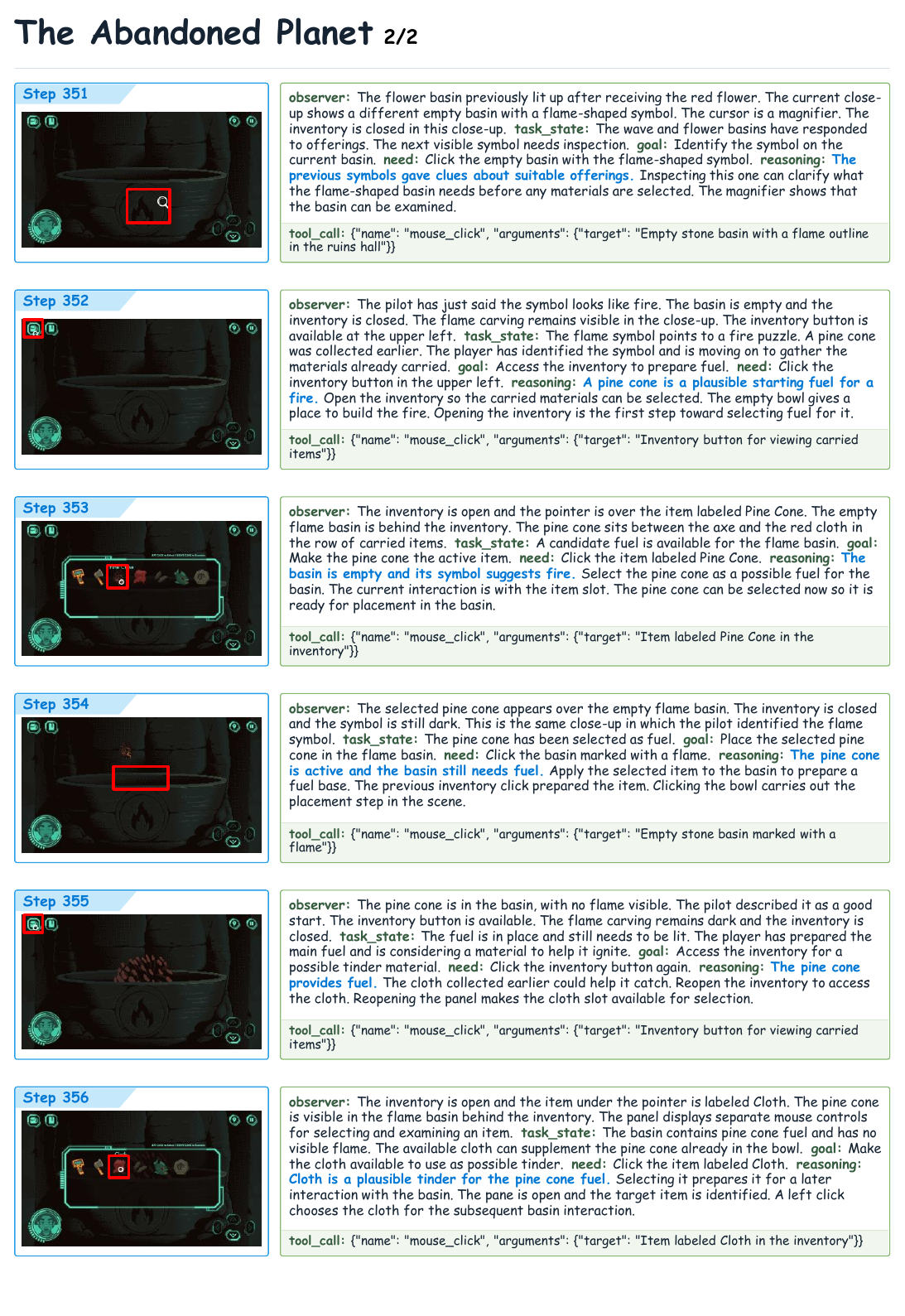}
  \caption{\textbf{\textit{The Abandoned Planet}, original steps 351--356.}}
  \label{fig:sup-planet-two} %
\end{figure*}

\clearpage

\twocolumn

{
    \small
    \bibliographystyle{ieeenat_fullname}
    \bibliography{main}
}

\end{document}